\documentclass[10pt,onecolumn]{IEEEtran}
\usepackage{amsthm}
\usepackage{times,amssymb,amsmath,amsfonts,float,nicefrac,color,bbm,mathrsfs,caption,float}
\usepackage{algorithm,enumerate,multirow,caption,tikz,graphicx}
\usepackage[mathscr]{eucal}
\usepackage{sidecap}
\usepackage{algpseudocode}
\usepackage{verbatim}
\usepackage{epstopdf}
\usepackage{textcomp}
\usetikzlibrary{shapes,arrows}
\usepackage{stmaryrd}
\usepackage{mathabx}
\usepackage[noadjust]{cite}
\usepackage{booktabs}
\usepackage[top=1in,bottom=1in,left=1in,right=1in]{geometry}
\allowdisplaybreaks

\newtheorem{theorem}{Theorem}[section]

\newtheorem{lemma}[theorem]{Lemma}
\newtheorem{proposition}[theorem]{Proposition}
\newtheorem{definition}[theorem]{Definition}

\newtheorem{cnstr}{Construction}

\newcounter{remark}[section]
\newenvironment{remark}[1][]{\refstepcounter{remark}\par\medskip
   \noindent \textbf{Remark~\thesection.\theremark. #1} \rmfamily}{\medskip}
   \newcounter{example}[section]
\newenvironment{example}[1][]{\refstepcounter{example}\par\medskip
   \noindent \textbf{Example~\thesection.\theexample. #1} \rmfamily}{\medskip}

\newcommand{\RN}[1]{%
  \textup{\expandafter{\romannumeral#1}}%
}

\newcommand\remove[1]{}
\newcommand{\nc}{\newcommand}

\allowdisplaybreaks

\def\mathbi#1{{\textbf{\textit #1}}}
\nc\bfa{{\boldsymbol a}}\nc\bfA{{\boldsymbol A}}\nc\cA{{\mathscr A}}\nc\sA{{\mathscr A}}
\nc\bfb{{\boldsymbol b}}\nc\bfB{{\boldsymbol B}}\nc\cB{{\mathscr B}}\nc\sB{{\mathscr B}}
\nc\bfc{{\boldsymbol c}}\nc\bfC{{\boldsymbol C}}\nc\cC{{\mathscr C}}\nc\sC{{\mathscr C}}
\nc\bfd{{\boldsymbol d}}\nc\bfD{{\boldsymbol D}}\nc\cD{{\mathscr D}}
\nc\bfe{{\boldsymbol e}}\nc\bfE{{\boldsymbol E}}\nc\cE{{\mathscr E}}
\nc\bff{{\boldsymbol f}}\nc\bfF{{\boldsymbol F}}\nc\cF{{\mathscr F}}\nc\sF{{\mathscr F}}
\nc\bfg{{\boldsymbol g}}\nc\bfG{{\boldsymbol G}}\nc\cG{{\mathscr G}}
\nc\bfh{{\boldsymbol h}}\nc\bfH{{\boldsymbol H}}\nc\cH{{\mathscr H}}
\nc\bfi{{\boldsymbol i}}\nc\bfI{{\boldsymbol I}}\nc\cI{{\mathscr I}}\nc\sI{{\mathscr I}}
\nc\bfj{{\boldsymbol j}}\nc\bfJ{{\boldsymbol J}}\nc\cJ{{\mathscr J}}
\nc\bfk{{\boldsymbol k}}\nc\bfK{{\boldsymbol K}}\nc\cK{{\mathscr K}}
\nc\bfl{{\boldsymbol l}}\nc\bfL{{\boldsymbol L}}\nc\cL{{\mathscr L}}
\nc\bfm{{\boldsymbol m}}\nc\bfM{{\boldsymbol M}}\nc\cM{{\mathscr M}}
\nc\bfn{{\boldsymbol n}}\nc\bfN{{\boldsymbol N}}\nc\cN{{\mathscr N}}
\nc\bfo{{\boldsymbol o}}\nc\bfO{{\boldsymbol O}}\nc\cO{{\mathscr O}}
\nc\bfp{{\boldsymbol p}}\nc\bfP{{\boldsymbol P}}\nc\cP{{\mathscr P}}\nc\eP{{\EuScriptP}}\nc\fP{{\mathfrak P}}
\nc\bfq{{\boldsymbol q}}\nc\bfQ{{\boldsymbol Q}}\nc\cQ{{\mathscr Q}}
\nc\bfr{{\boldsymbol r}}\nc\bfR{{\boldsymbol R}}\nc\cR{{\mathscr R}}\nc\sR{{\mathscr R}}
\nc\bfs{{\boldsymbol s}}\nc\bfS{{\boldsymbol S}}\nc\cS{{\mathscr S}}
\nc\bft{{\boldsymbol t}}\nc\bfT{{\boldsymbol T}}\nc\cT{{\mathscr T}}
\nc\bfu{{\boldsymbol u}}\nc\bfU{{\boldsymbol U}}\nc\cU{{\mathscr U}}
\nc\bfv{{\boldsymbol v}}\nc\bfV{{\boldsymbol V}}\nc\cV{{\mathscr V}}\nc\sV{{\mathscr V}}
\nc\bfw{{\boldsymbol w}}\nc\bfW{{\boldsymbol W}}\nc\cW{{\mathscr W}}\nc\sW{{\mathscr W}}
\nc\bfx{{\boldsymbol x}}\nc\bfX{{\boldsymbol X}}\nc\cX{{\mathscr X}}
\nc\bfy{{\boldsymbol y}}\nc\bfY{{\boldsymbol Y}}\nc\cY{{\mathscr Y}}
\nc\bfz{{\boldsymbol z}}\nc\bfZ{{\boldsymbol Z}}\nc\cZ{{\mathscr Z}}

\DeclareMathOperator{\sign}{sign}
\DeclareMathOperator{\TV}{TV}
\DeclareMathOperator{\Var}{Var}

\DeclareMathOperator{\kl}{kl}
\DeclareMathOperator{\RAP}{RAP}
\DeclareMathOperator{\RR}{RR}
\DeclareMathOperator{\OPT}{OPT}

\begin{document}

%\pagenumbering{gobble}

\title{Optimal Schemes for Discrete Distribution Estimation under Locally Differential Privacy}

\author{\IEEEauthorblockN{Min Ye} \hspace*{1in}
\and \IEEEauthorblockN{Alexander Barg}}

\maketitle
{\renewcommand{\thefootnote}{}\footnotetext{

\vspace{-.2in}
 
\noindent\rule{1.5in}{.4pt}

The authors are with Dept. of ECE and ISR, University of Maryland, College Park, MD 20742. Emails: yeemmi@gmail.com and abarg@umd.edu. Research supported by NSF grants CCF1422955 and CCF1618603.}
}
\renewcommand{\thefootnote}{\arabic{footnote}}
\setcounter{footnote}{0}

\begin{abstract}
We consider the minimax estimation problem of a discrete distribution with support size $k$ under privacy constraints. 
A privatization scheme is applied to each raw sample independently, and we need to estimate the distribution of the raw samples from the privatized samples. A positive number $\epsilon$ measures the privacy level of a privatization scheme. 
For a given $\epsilon,$ we consider the problem of constructing optimal privatization schemes with $\epsilon$-privacy level, i.e., 
schemes that minimize the expected estimation loss for the worst-case distribution.
Two schemes in the literature provide order optimal performance in the high privacy regime where $\epsilon$ is very close to $0,$ and in the low privacy regime where $e^{\epsilon}\approx k,$ respectively.

In this paper, we propose a new family of schemes which substantially improve the performance of the existing schemes in
the medium privacy regime when $1\ll e^{\epsilon} \ll k.$ More concretely, we prove that when $3.8 < \epsilon <\ln(k/9) ,$ our schemes 
reduce the expected estimation loss by $50\%$ under $\ell_2^2$ metric and by $30\%$ under $\ell_1$ metric over the existing schemes. We also prove a lower bound for the region $e^{\epsilon} \ll k,$ which implies that our schemes are order optimal in this regime.
\end{abstract}

\section{introduction}
A major challenge in the statistical analysis of user data is the conflict between learning accurate statistics and protecting sensitive information about the individuals. To study this tradeoff, we need a formal definition of privacy, and {\em differential privacy} has been put forth as one such candidate \cite{Dwork06, Dwork08}.
Roughly speaking, differential privacy requires that the adversary not be able to reliably infer an individual's data from public statistics even with access to all the other users' data.
The concept of differential privacy has been developed in two different contexts: the {\em global privacy}
context (for instance, when institutions release statistics of groups of people) \cite{Ghosh12}, and the {\em local privacy} context when individuals disclose their personal data \cite{Duchi13}.

In this paper, we consider the minimax estimation problem of a discrete distribution with support size $k$ under locally differential privacy.
This problem has been studied in the non-private setting \cite{Kamath15, Lehmann06}, where we can learn the distribution from the raw samples.
In the private setting, we need to estimate the distribution of raw samples from the privatized samples, which are generated independently from each raw sample according to a conditional distribution (also called privatization scheme) $\mathbi{Q}.$
Given a privacy parameter $\epsilon>0,$
we say that $\mathbi{Q}$ is $\epsilon$-locally differentially private if the probabilities of the same output conditional on different inputs differ by a factor of at most $e^{\epsilon}.$ Clearly, smaller $\epsilon$ means that it is more difficult to infer the original data from the privatized samples, and thus leads to higher privacy.
For a given $\epsilon,$ our objective is to find the optimal privatization scheme with $\epsilon$-privacy level to minimize the expected estimation loss for the worst case distribution.
In this paper, we are mainly
concerned with the scenario where we have a large number of
samples, which captures the modern trend toward ``big data" analytics.

\subsection{\em Existing results:} The following two privatization schemes are the most well-known in the literature: the $k$-ary Randomized Aggregatable Privacy-Preserving Ordinal Response ($k$-RAPPOR) scheme \cite{Duchi13a, Erlingsson14}, and the $k$-ary Randomized Response ($k$-RR) scheme
\cite{Warner65,Kairouz14,Pastore16}.
The $k$-RAPPOR scheme is order optimal in the high privacy regime where $\epsilon$ is very close to $0,$
and the $k$-RR scheme is order optimal in the low privacy regime where $e^{\epsilon} \approx k$ \cite{Kairouz16}. At the
same time, to the best of our knowledge, no schemes work well in the medium privacy regime, where $e^{\epsilon}$ is far from both $1$ or $k.$ 
Arguably, this regime is of practical importance: Indeed, if $\epsilon$ is too close to $0,$ then we may need too many samples to estimate the distribution accurately; on the other hand, taking $\epsilon$ too large can compromise the privacy requirement.

Duchi et al.~\cite{Duchi16} gave a tight lower bound on the minimax private estimation loss for the high privacy regime where $\epsilon$ is very close to $0$. At the same time, no meaningful lower bounds are known for the medium privacy regime.

\subsection{\em Our contributions:} In this paper we first propose a family of new privatization schemes which are order-optimal in the medium to high privacy regimes when $e^{\epsilon} \ll k.$ We show that our schemes are better than the two existing schemes in the medium privacy regime where 
$1\ll e^\epsilon \ll k.$ 
For instance, we show that for the $\ell_2^2$ loss our scheme outperforms the $k$-RR scheme by a factor of $\Theta(k/e^\epsilon)$,
and prove similar results for $k$-RAPPOR and $\ell_1$ loss.
We also show  that 
when $3.8 < \epsilon <\ln(k/9) ,$ our schemes reduce the expected estimation loss by $50\%$ under the $\ell_2^2$ metric and by $30\%$ under the $\ell_1$ metric over the existing schemes. This compares favorably with the existing literature (e.g.,~\cite{Kairouz16}) where the improvement of several percentage points constitutes a substantial advance.
Second, we prove a tight lower bound for the whole region $e^{\epsilon} \ll k,$ which implies that our schemes are order optimal in this regime.  
We also prove that in order to obtain the optimal performance, we only need to consider the privatization schemes formed by {\em extremal configurations}, namely, we can restrict ourselves to the privatization schemes
with finite output alphabet and the property that the ratio between the probabilities of a given output conditional on different inputs is either $1$ or $e^{\epsilon}.$

After this paper was completed, we learned that the privatization scheme and the empirical estimator that we derive have been proposed earlier in the work of Wang et al. \cite{Wang16} under the name of $k$-subset mechanisms. The authors of \cite{Wang16} showed that the $k$-subset mechanisms outperform the $k$-RR and $k$ RAPPOR schemes, quantifying the improvement in experimental results. They also proposed the efficient implementation of their estimator that we discuss in Remark~\ref{Sect:new}.\ref{rem:implement} below. At the same time, \cite{Wang16} does not include a detailed analysis of the existing schemes and the new proposal in the medium privacy regime. Finally,
\cite{Wang16} does not address lower bounds on the risk and therefore does not include the statement that the proposed privatization mechanisms are order-optimal in terms of the expected estimation loss. 

Our paper is organized as follows: in Sect.~\ref{Sect:pre} we formulate the problem and give necessary background. In Sect.~\ref{Sect:new} we introduce our new schemes and evaluate their performance.
In Sect.~\ref{Sect:lb} we prove the optimality of extremal configuration and the tight lower bound.

\section{preliminaries and problem formulation}\label{Sect:pre}
\textbf{Notation:}
Let $\cX=\{1,2,\dots,k\}$ be the source alphabet and let $\mathbi{p}=(p_1,p_2,\dots,p_k)$ be a probability distribution on $\cX.$
Denote by $\Delta_k=\{\mathbi{p}\in \mathbb{R}^k: p_i\ge 0 \text{~for~} i=1,2,\dots,k, \sum_{i=1}^k p_i=1\}$ the $k$-dimensional probability simplex. Let $X$ be a random variable (RV) that takes values on $\cX$ according to $\mathbi{p}$, so that $p_i=\mathbi{p}(X=i).$ Denote by $X^n=(X^{(1)},X^{(2)},\dots,X^{(n)})$ the vector formed of $n$ independent copies of the RV $X.$

In the classical (non-private) distribution estimation problem, we are given direct access to i.i.d. samples
$\{X^{(i)}\}_{i=1}^n$ drawn according to some unknown distribution $\mathbi{p}\in \Delta_k.$ Our goal is to estimate $\mathbi{p}$ based on the samples \cite{Lehmann06}. We define an estimator $\hat{\mathbi{p}}$ as a function
$\hat{\mathbi{p}}:\cX^n \to \mathbb{R}^k,$ and assess the quality of the estimator $\hat{\mathbi{p}}$ in terms of the risk (expected loss)
$$
\underset{X^n\sim \mathbi{p}^n}{\mathbb{E}} \ell(\hat{\mathbi{p}}(X^n), \mathbi{p}),
$$
where $\ell$ is some loss function. The minimax risk is defined as the following saddlepoint problem:
$$
r_{k,n}^{\ell}:= \inf_{\hat{\mathbi{p}}} \sup_{\mathbi{p}\in \Delta_k} 
\underset{X^n\sim \mathbi{p}^n}{\mathbb{E}} \ell(\hat{\mathbi{p}}(X^n), \mathbi{p}).
$$

In the private distribution estimation problem, we can no longer access the raw samples $\{X^{(i)}\}_{i=1}^n.$ Instead, we estimate the distribution $\mathbi{p}$ from the privatized samples $\{Y^{(i)}\}_{i=1}^n,$ obtained by applying a privatization mechanism $\mathbi{Q}$ independently to each raw sample $X^{(i)}.$ A {\em privatization mechanism} (also called privatization scheme) $\mathbi{Q}:\cX\to\cY$ is simply a conditional distribution $\mathbi{Q}_{Y|X}.$ The
privatized samples $Y^{(i)}$ take values in a set $\cY$ (the ``output alphabet'') that does not have to be the same as $\cX.$

The quantities $\{Y^{(i)}\}_{i=1}^n$ are i.i.d. samples drawn according to the marginal distribution $\mathbi{m}$ given by
\begin{equation}\label{eq:defm}
\mathbi{m}(S)=\sum_{i=1}^k \mathbi{Q}(S|i)p_i
\end{equation}
 for any $S\in \sigma(\cY),$ where $\sigma(\cY)$ denotes an appropriate $\sigma$-algebra on $\cY.$
In accordance with this setting, the estimator $\hat{\mathbi{p}}$ is a measurable function $\hat{\mathbi{p}}:\cY^n\to \mathbb{R}^k.$
Define the {\em minimax risk} of the privatization mechanism $\mathbi{Q}$ as
$$
r_{k,n}^{\ell} (\mathbi{Q}):= \inf_{\hat{\mathbi{p}}} \sup_{\mathbi{p}\in \Delta_k} 
\underset{Y^n\sim \mathbi{m}^n}{\mathbb{E}} \ell(\hat{\mathbi{p}}(Y^n), \mathbi{p}),
$$
where $\mathbi{m}^n$ is the $n$-fold product distribution and $\mathbi m$ is given by \eqref{eq:defm}.
\begin{definition}
For a given $\epsilon>0,$
a privatization mechanism $\mathbi{Q}:\cX\to\cY$ is said to be {\em $\epsilon$-locally differentially private}
\footnote{Following the existing literature, we use the quantity $e^\epsilon$ as the measure of privacy level even though $\epsilon$
is never used separately in our derivations and results.} if 
\begin{equation}\label{eq:defep}
\sup_{S\in\sigma(\cY)} \frac{\mathbi{Q}(Y\in S|X=x)}{\mathbi{Q}(Y\in S|X=x')} \le e^{\epsilon}
\text{~for all~} x,x'\in\cX.
\end{equation}
\end{definition}

Denote by $\cD_{\epsilon}$ the set of all $\epsilon$-locally differentially private mechanisms. Given a privacy level 
$\epsilon,$ we want to find the optimal $\mathbi{Q}\in\cD_{\epsilon}$ with the smallest possible minimax risk among all the 
$\epsilon$-locally differentially private mechanisms. We further define the $\epsilon$-private minimax risk as
 \begin{equation}
r_{\epsilon,k,n}^{\ell} := \inf_{\mathbi{Q}\in\cD_{\epsilon}}r_{k,n}^{\ell} (\mathbi{Q}).
  \label{eq:rekn}
  \end{equation}
In Sect.~\ref{Sect:lb}, we show that it suffices to restrict oneself to finite output alphabet $\cY,$ i.e.,
$$
r_{\epsilon,k,n}^{\ell} = \inf_{\mathbi{Q}\in\cD_{\epsilon,F}}r_{k,n}^{\ell} (\mathbi{Q}),
$$
where $\cD_{\epsilon,F}$ is the set of $\epsilon$-locally differentially private mechanisms with finite output alphabet. 
%Thus we only concern with private mechanisms with finite output alphabet in this paper.
For $\mathbi{Q}\in \cD_{\epsilon,F},$ Eq.~\eqref{eq:defep} is equivalent to
$$
\frac{\mathbi{Q}(Y=y|X=x)}{\mathbi{Q}(Y=y|X=x')} \le e^{\epsilon}
\text{~for all~} x,x'\in\cX \text{~and~} y\in\cY.
$$
We shall also write the definition of the marginal distribution $\mathbi{m}$  in \eqref{eq:defm}  as $\mathbi{m}=\mathbi{p}\mathbi{Q}.$

We will use standard distance functions on distributions defined on finite sets $\cY.$ The {\em KL divergence} between
two such distributions $\mathbi{m}_1$ and $\mathbi{m}_2$ is defined as
$$
D_{\kl}(\mathbi{m}_1 || \mathbi{m}_2) := \sum_{y\in\cY}  \mathbi{m}_1(y) \log \frac{\mathbi{m}_1(y)}
{\mathbi{m}_2(y)}.
$$
The {\em total variation distance} between $\mathbi{m}_1$ and $\mathbi{m}_2$ is defined as
$$
\| \mathbi{m}_1 - \mathbi{m}_2 \|_{\TV} := \max_{A\subseteq \cY} |\mathbi{m}_1(A) - \mathbi{m}_2(A)|
=\frac{1}{2}\sum_{y\in\cY} |\mathbi{m}_1(y) - \mathbi{m}_2(y)|.
$$

\section{new schemes} \label{Sect:new}
In this section we introduce a family of new privatization schemes.
Our schemes are parameterized by the integer $d\in\{1,2,\dots,k-1\}.$ Given $d,$ let the output alphabet be 
$\cY_{k,d}=\{y\in \{0,1\}^k: \sum_{i=1}^k y_i=d\}.$ Clearly, $|\cY_{k,d}|=\binom{k}{d}.$ Define
   \begin{equation}\label{eq:defQ}
\mathbi{Q}_{k,d}(y|i)=\frac{e^\epsilon y_i+(1-y_i)}{\binom{k-1}{d-1}e^{\epsilon}+\binom{k-1}{d}} %(e^\epsilon y_i+(1-y_i))
%\left\{\begin{array}{cc}e^{\epsilon} & \text{if }y_i=1\\1 & \text{if }y_i=0\end{array}\right. ,
    \end{equation}
for all $y\in\cY_{k,d}$ and all $i\in\cX.$
To define the estimator for $\mathbi{Q}_{k,d},$ we need to calculate the marginal distribution of each coordinate of the output. 
We begin with a concrete example to illustrate the method of the calculation.
\begin{example} Let $\cY_{4,2}\subset\{0,1\}^4$ be the set of all vectors with two ones and two zeros.
%$\cY_{4,2}=\{(1,1,0,0), (1,0,1,0), (1,0,0,1), (0,1,1,0), (0,1,0,1), (0,0,1,1)\}.$ 
For any $i=1,\dots,4,$  $Y_i$ is a Bernoulli random variable. Consider the event $A_1:=\{Y_1=1\}=\{Y\in\{(1,1,0,0), (1,0,1,0), (1,0,0,1)\}\}.$
We have
  $$
     Q(A_1|X=1)=\frac{3 e^{\epsilon}}{3 e^{\epsilon}+3},\;\;
     Q(A_1|X=i)=\frac{e^{\epsilon} + 2}{3 e^{\epsilon}+3} \quad\text{ for $i=2,3,4.$}
     $$
 Using  \eqref{eq:defm}, we obtain
 $$
 \mathbi{m}_{4,2}(Y_1=1)%\frac{3 e^{\epsilon}p_1 + (e^{\epsilon} + 2)(p_2+p_3+p_4)}{3 e^{\epsilon}+3}
=\frac{3 e^{\epsilon}p_1 + (e^{\epsilon} + 2)(1-p_1)}{3 e^{\epsilon}+3},
$$ 
where $\mathbi{m}_{4,2}
=\mathbi{p}\mathbi{Q}_{4,2}.$
\end{example}

For $d>1,$ we can derive the marginal distribution of each coordinate of the output using the method illustrated above:
  \begin{align}
q_i=\mathbi{m}_{k,d}(Y_i=1)&=
\frac{\binom{k-1}{d-1}e^{\epsilon}p_i+(\binom{k-2}{d-2}e^{\epsilon}+\binom{k-2}{d-1})(1-p_i)}
{\binom{k-1}{d-1}e^{\epsilon}+\binom{k-1}{d}}\nonumber\\
&=\frac{(k-1)e^{\epsilon}p_i+((d-1)e^{\epsilon}+k-d)(1-p_i)}
{(k-1)e^{\epsilon}+\frac{(k-1)(k-d)}{d}}\nonumber\\
&=\frac{(k-d)(e^{\epsilon}-1)p_i+(d-1)e^{\epsilon}+k-d}
{(k-1)e^{\epsilon}+\frac{(k-1)(k-d)}{d}},\label{eq:Py}
  \end{align}
where $\mathbi{m}_{k,d}=\mathbi{p}\mathbi{Q}_{k,d}.$
It is easy to check that the final expression in \eqref{eq:Py} also holds for $d=1.$

Solving for $p_i$, we obtain the empirical estimator of $\mathbi{p}$ under $\mathbi{Q}_{k,d}$ in the following form
\begin{equation}\label{eq:emp}
\hat{p_i}=\left(\frac{(k-1)e^{\epsilon}+\frac{(k-1)(k-d)}{d}}{(k-d)(e^{\epsilon}-1)}\right)\frac{T_i}{n}
-\frac{(d-1)e^{\epsilon}+k-d}{(k-d)(e^{\epsilon}-1)},
\end{equation}
where $T_i=\sum_{j=1}^n Y_i^{(j)}.$

\begin{remark}\label{rem:implement}
When $d$ is large, the denominator in \eqref{eq:defQ} is exponentially large in $k.$ In practice, $k$ can be several hundred to several thousand, and the conditional probability of each output can thus be very small. To circumvent computational difficulties in \eqref{eq:defQ}, we suggest the following recursive scheme for 
implementing
$\mathbi{Q}_{k,d}.$ Given a raw sample (input) $i\in\cX,$ we first produce the $i$-th coordinate of the privatized sample (output) $Y_i$ according to the distribution:
\begin{gather*}
\mathbi{Q}_{k,d}(Y_i=1|i) = \frac{\binom{k-1}{d-1}e^{\epsilon}}{\binom{k-1}{d-1}e^{\epsilon}
+\binom{k-1}{d}} = \frac{d e^{\epsilon}}{d e^{\epsilon}+k-d}, \\
\mathbi{Q}_{k,d}(Y_i=0|i) =  \frac{k-d}{d e^{\epsilon}+k-d}.
\end{gather*}
If $Y_i$ is $1,$ then we choose $d-1$ distinct elements $\{i_1,i_2,\dots,i_{d-1}\}$ uniformly from $\cX\setminus \{i\},$ and set $Y_j=1$ if $j\in\{i,i_1,i_2,\dots,i_{d-1}\}$ and $Y_j=0$ otherwise.
If $Y_i$ is $0,$ then we choose $d$ distinct elements $\{i_1,i_2,\dots,i_d\}$ uniformly from $\cX\setminus \{i\},$ and set $Y_j=1$ if $j\in\{i_1,i_2,\dots,i_d\}$ and $Y_j=0$ otherwise.
When we choose $d$ distinct elements uniformly from the set $\cX\setminus\{i\},$ we choose them one by one: we first choose $i_1$ uniformly from $\cX\setminus\{i\},$ then we choose $i_2$ uniformly from $\cX\setminus\{i,i_1\},$ so on and so forth, until we choose $d$ elements. It is easy to verify that the procedure we described above produces exactly the same distribution as designed in \eqref{eq:defQ}.  Moreover, the smallest probability we need to deal with is at least $1/k$ in this procedure. So the scheme $\mathbi{Q}_{k,d}$ can be efficiently implemented in practice.
\end{remark}

Let us calculate the risk under the $\ell_2^2$ loss and $\ell_1$ loss.
\begin{proposition}\label{prop:risks} Suppose that the privatization scheme is 
 $\mathbi{Q}_{k,d}$ and the empirical estimator is given by \eqref{eq:emp}.
Let $\mathbi{m}=\mathbi{p}\mathbi{Q}_{k,d}.$
For all $\epsilon, n$ and $k,$ we have that
    \begin{equation}\label{eq:L2}
    \underset{Y^n\sim \mathbi{m}^n}{\mathbb{E}} \ell_2^2(\hat{\mathbi{p}}(Y^n),\mathbi{p})=\frac{1}{n}
     \Big(\frac{(d(k-2)+1)e^{2\epsilon}}{(k-d)(e^{\epsilon}-1)^2} +\frac{2(k-2)e^{\epsilon}}{(e^{\epsilon}-1)^2}
     +\frac{(k-2)(k-d)+1}{d(e^{\epsilon}-1)^2} -\sum_{i=1}^k p_i^2 \Big).
    \end{equation}
The expected $\ell_1$ loss in the limit of large $n$ is given by
\begin{equation}\label{eq:L1}
\underset{Y^n\sim \mathbi{m}^n}{\mathbb{E}} \ell_1(\hat{\mathbi{p}}(Y^n),\mathbi{p}) =
\frac{1}{e^{\epsilon}-1} \sum_{i=1}^k \sqrt{\frac{2}{\pi n} \Big( (e^{\epsilon}-1)p_i+ \frac{(d-1)e^{\epsilon}}{k-d}+1 \Big)
\Big((e^{\epsilon}-1)(1-p_i)+\frac{k-1}{d} \Big)} + o\Big(\frac 1{\sqrt n}\Big).
\end{equation}
%where $\mathbi{m}=\mathbi{p}\mathbi{Q}_{k,d}.$
\end{proposition}
The proof is elementary but somewhat tedious. It is given in Appendix \ref{ap:risk}.

Next we find the optimal value of $d$ to minimize the $\ell_2^2$ risk and $\ell_1$ risk for the worst case distribution.
\begin{proposition} Let  $\epsilon$ be a given privacy level. The optimal choice of $d$ for both the $\ell_2^2$ risk and the $\ell_1$ risk is given by either
$d=\lceil k/(e^{\epsilon}+1) \rceil$ or $d=\lfloor k/(e^{\epsilon}+1)\rfloor.$
\end{proposition}
\begin{IEEEproof}
Let us begin with the $\ell_2^2$ case. Starting from \eqref{eq:L2}, we need to minimize the terms that contain $d:$
\begin{align*}
 \frac{(d(k-2)+1)e^{2\epsilon}}{(k-d)(e^{\epsilon}-1)^2} 
+\frac{(k-2)(k-d)+1}{d(e^{\epsilon}-1)^2} =\frac{1}{(e^{\epsilon}-1)^2}
\Big( (k-2)\Big(\frac{d}{k-d}e^{2\epsilon}+\frac{k-d}{d}\Big) +\frac{e^{2\epsilon}}{k-d}+\frac{1}{d} \Big).
\end{align*}
Denote the expression in the outer parentheses on the right-hand side by $g(d).$
We have
$$
g'(d) = (k-1)^2 \Big( \frac{e^{2\epsilon}}{(k-d)^2} - \frac{1}{d^2} \Big).
$$
It is easy to see that $g'(d)$ is an increasing function in the interval $d\in(0,k).$ Thus the minimum of $g(d)$ occurs when $g'(d)=0,$ namely, when $d=k/(e^{\epsilon}+1).$ Notice that $0 < k/(e^{\epsilon}+1) \le k/2.$
Since $d$ is an integer between $1$ and $k,$ the minimum is attained at one of the nearest integers to $k/(e^{\epsilon}+1).$

As for the $\ell_1$ loss,
by using the Cauchy-Schwarz inequality twice, we can easily see that the right-hand side of \eqref{eq:L1} reaches maximum for the uniform distribution $\mathbi{p}_U=(1/k,1/k,\dots,1/k):$
\begin{equation}\label{eq:maxU}
\underset{Y^n\sim \mathbi{m}_U^n}{\mathbb{E}} \ell_1(\hat{\mathbi{p}}(Y^n),\mathbi{p}_U) =
 \frac{1}{e^{\epsilon}-1}\sqrt{\frac{2(k-1)}{\pi n}
 \Big( e^{\epsilon}-1+\frac{k(d-1)e^{\epsilon}}{k-d}+k \Big)
\Big(e^{\epsilon}+\frac{k-d}{d} \Big)} +o\Big(\frac 1{\sqrt n}\Big),
\end{equation}
where $\mathbi{m}_U=\mathbi{p}_U\mathbi{Q}_{k,d}.$
Let $$
     f(d)=\Big( e^{\epsilon}-1+\frac{k(d-1)e^{\epsilon}}{k-d}+k \Big)
\Big(e^{\epsilon}+\frac{k-d}{d} \Big)
   $$
 We find 
$f'(d)=k(k-1)\big( \frac{e^{2\epsilon}}{(k-d)^2} - \frac{1}{d^2} \big).$
It is easy to see that $f'(d)$ is an increasing function in the interval $d\in(0,k).$ Thus the minimum of $f(d)$ occurs when $f'(d)=0,$ namely, when $d=k/(e^{\epsilon}+1).$
Since $d\in\{1,\dots,k\}$ is integer, this concludes the proof. 
\end{IEEEproof}

In order to avoid the case $d=0,$ below we take $d=\lceil k/(e^{\epsilon}+1) \rceil$ as a convenient and nearly optimal choice.
The next proposition gives upper bounds on the $\ell_2^2$ risk and $\ell_1$ risk for this value of $d$.

\begin{proposition} \label{Prop:ub}
Let $k\ge \max(4,e^{\epsilon}+1).$ Suppose that the privatization scheme is 
 $\mathbi{Q}_{k,d}, d=\lceil k/(e^{\epsilon}+1) \rceil$ and the corresponding empirical estimator is given by \eqref{eq:emp}.
 Let $\mathbi{m}=\mathbi{p}\mathbi{Q}_{k,d}.$
For all $\epsilon, n,k$ and $\mathbi{p}\in\Delta_k,$ we have that
\begin{equation}\label{eq:longub}
\underset{Y^n\sim \mathbi{m}^n}{\mathbb{E}} \ell_2^2(\hat{\mathbi{p}}(Y^n),\mathbi{p})
< \frac{4k e^{\epsilon}}{n (e^{\epsilon}-1)^2}
\Big(1 + \frac{2e^{\epsilon}+3}{4k}\Big),
\end{equation}
and for large $n,$ we have that 
\begin{equation}\label{eq:l1ub}
\underset{Y^n\sim \mathbi{m}^n}{\mathbb{E}} \ell_1(\hat{\mathbi{p}}(Y^n),\mathbi{p}) <
\sqrt{\frac{8e^{\epsilon}}{\pi n }}  \frac{k }{(e^{\epsilon}-1) } 
\Big (1+\frac{e^{\epsilon}+1}{4k}\Big),
\end{equation}

In the regime $e^{\epsilon} \ll k$ we have
\begin{equation}\label{eq:orderub}
\underset{Y^n\sim \mathbi{m}^n}{\mathbb{E}} \ell_2^2(\hat{\mathbi{p}}(Y^n),\mathbi{p}) =
\Theta\Big(\frac{k e^{\epsilon}}{n (e^{\epsilon}-1)^2}\Big), \quad
\underset{Y^n\sim \mathbi{m}^n}{\mathbb{E}} \ell_1(\hat{\mathbi{p}}(Y^n),\mathbi{p})  =
\Theta\Big(\frac{k }{e^{\epsilon}-1} \sqrt{\frac{e^\epsilon}{n}}\Big ).
\end{equation}
In the regime $1\ll e^{\epsilon} \ll k$ we have
\begin{equation}\label{eq:orderub-1}
\underset{Y^n\sim \mathbi{m}^n}{\mathbb{E}} \ell_2^2(\hat{\mathbi{p}}(Y^n),\mathbi{p}) =
 \Theta\Big(\frac{k}{n e^{\epsilon}}\Big) , \quad
\underset{Y^n\sim \mathbi{m}^n}{\mathbb{E}} \ell_1(\hat{\mathbi{p}}(Y^n),\mathbi{p}) =
 \Theta\Big(\frac{k}{\sqrt{ne^{\epsilon}}}\Big).
\end{equation}
\end{proposition}
\begin{IEEEproof}
We begin with proving the upper bound on $\ell_2^2$ risk.
We know that $k/(e^{\epsilon}+1) \le d = \lceil k/(e^{\epsilon}+1) \rceil \le k/(e^{\epsilon}+1) +1.$
In \eqref{eq:L2}, there are only two terms containing $d.$ The first one
$\frac{(d(k-2)+1)e^{2\epsilon}}{(k-d)(e^{\epsilon}-1)^2}$ is an increasing function of $d$ for $d\in(0,k),$
so replacing $d = \lceil k/(e^{\epsilon}+1) \rceil$ with $k/(e^{\epsilon}+1) +1$ gives an upper bound on this term:
\begin{align}
\frac{(d(k-2)+1)e^{2\epsilon}}{(k-d)(e^{\epsilon}-1)^2}
& \le  \frac{((\frac{k}{e^{\epsilon}+1}+1)(k-2)+1)e^{2\epsilon}}{(\frac{e^{\epsilon}k}{e^{\epsilon}+1} - 1)(e^{\epsilon}-1)^2}\nonumber\\
& =  \frac{ (k-2)e^{\epsilon}}
{ (e^{\epsilon}-1)^2} \Big(1+\frac{e^{\epsilon}+1}{k}\Big) 
 \Big(1+\frac{1}{(\frac{k}{e^{\epsilon}+1}+1)(k-2)}\Big) \Big(1- \frac{e^{\epsilon}+1}{e^{\epsilon}k}\Big)^{-1} \nonumber\\
& <  \frac{ (k-2)e^{\epsilon}}
{ (e^{\epsilon}-1)^2} \Big(1+\frac{e^{\epsilon}+1}{k}\Big) 
 \Big(1+\frac{e^{\epsilon}+1}{k(k-2)}\Big) \Big(1- \frac{2}{k}\Big)^{-1} \nonumber\\
& \overset{(a)}{\le}   \frac{ (k-2)e^{\epsilon}}
{ (e^{\epsilon}-1)^2} \Big(1+\frac{2(e^{\epsilon}+1)}{k}\Big) 
 \Big(1+ \frac{4}{k}\Big) \nonumber\\
& <  \frac{ k e^{\epsilon}}
{ (e^{\epsilon}-1)^2} \Big(1+\frac{2(e^{\epsilon}+1)}{k}\Big)  \Big(1+ \frac{2}{k}\Big) \nonumber\\
& \overset{(b)}{\le}   \frac{ k e^{\epsilon}}
{ (e^{\epsilon}-1)^2} \Big(1+\frac{2(e^{\epsilon}+4)}{k}\Big) ,\label{eq:ub1}
\end{align}
where $(a)$ follows from the assumption that $k\ge \max(4, e^{\epsilon}+1)$
and the obvious inequality $(1-x)^{-1}\le 1+2x$ for all $x\in [0,1/2],$ and $(b)$ follows from the assumption that $k\ge e^{\epsilon}+1.$

 The second term in \eqref{eq:L2} that we need to analyze is $\frac{(k-2)(k-d)+1}{d(e^{\epsilon}-1)^2}.$ It is a decreasing function of $d$ for $d\in(0,k),$ so replacing $d = \lceil k/(e^{\epsilon}+1) \rceil$ with $k/(e^{\epsilon}+1)$ gives an upper bound on this term:
\begin{align}
\frac{(k-2)(k-d)+1}{d(e^{\epsilon}-1)^2}
& \le \frac{(k-2)e^{\epsilon}}{(e^{\epsilon}-1)^2}
\Big(1 + \frac{e^{\epsilon}+1}{k(k-2)e^{\epsilon}}\Big) \le 
\frac{(k-2)e^{\epsilon}}{(e^{\epsilon}-1)^2}
\Big(1 + \frac{2}{k(k-2)}\Big) \nonumber\\
& \le \frac{(k-2)e^{\epsilon}}{(e^{\epsilon}-1)^2}
\Big(1 + \frac{1}{k}\Big)
\le  \frac{k e^{\epsilon}}{(e^{\epsilon}-1)^2}
\Big(1 - \frac{1}{k}\Big).\label{eq:ub2}
\end{align}
Substituting inequalities \eqref{eq:ub1},\eqref{eq:ub2} into \eqref{eq:L2} and discarding the negative term
$(-\sum_{i=1}^k p_i^2),$ we obtain \eqref{eq:longub}.

Next we prove the upper bound on $\ell_1$ risk.
As we noted before, the right-hand side of \eqref{eq:L1} is maximum when $\mathbi{p}=\mathbi{p}_U$ 
where $\mathbi{p}_U=(1/k,1/k,\dots,1/k)$ is the uniform distribution. For this reason, we will bound from above the 
right-hand side of \eqref{eq:maxU}. Again there are only two terms in \eqref{eq:maxU} that contain $d.$ The first one is $\frac{k(d-1)e^{\epsilon}}{k-d}$ and it is an increasing function of $d$ for $d\in(0,k).$
Replacing $d = \lceil k/(e^{\epsilon}+1) \rceil$ with $k/(e^{\epsilon}+1) +1,$ we obtain the following upper bound on this term:
\begin{equation}\label{eq:ubl11}
\frac{k(d-1)e^{\epsilon}}{k-d}
\le k\Big(1-\frac{e^{\epsilon}+1}{k e^{\epsilon}}\Big)^{-1} \le k\Big(1-\frac{2}{k}\Big)^{-1}
\le k\Big(1+\frac{4}{k}\Big).
\end{equation}
The other term in  \eqref{eq:maxU} that involves $d$ is $\frac{k-d}{d}$ and it is a decreasing function of $d$ for $d\in(0,k).$ Replacing $d = \lceil k/(e^{\epsilon}+1) \rceil$ with $k/(e^{\epsilon}+1),$ we obtain the following upper bound on this term:
\begin{equation}\label{eq:ubl12}
\frac{k-d}{d} \le e^{\epsilon}.
\end{equation}
Substituting \eqref{eq:ubl11} and\eqref{eq:ubl12} into \eqref{eq:maxU}, we obtain the following inequality:
  \begin{align*}
\underset{Y^n\sim \mathbi{m}^n}{\mathbb{E}} \ell_1(\hat{\mathbi{p}}(Y^n),\mathbi{p}) 
& \le \frac{1}{e^{\epsilon}-1} 
\sqrt{2e^{\epsilon} \Big( e^{\epsilon}-1+k(1+\frac{4}{k})+k \Big)
} \sqrt{\frac{2k}{\pi n}} \Big(1-\frac{1}{k}\Big)^{1/2} \\
& =  \sqrt{\frac{8e^{\epsilon}}{\pi n }}  \frac{k }{(e^{\epsilon}-1) } 
\Big(1-\frac{1}{k}\Big)^{1/2}  \Big(1+\frac{e^{\epsilon}+3}{2k}\Big)^{1/2} \\
& \overset{(a)}{\le} \sqrt{\frac{8e^{\epsilon}}{\pi n }}  \frac{k }{(e^{\epsilon}-1) } 
\Big(1-\frac{1}{2k}\Big)  \Big(1+\frac{e^{\epsilon}+3}{4k}\Big) \\
& < \sqrt{\frac{8e^{\epsilon}}{\pi n }}  \frac{k }{(e^{\epsilon}-1) } 
\Big(1+\frac{e^{\epsilon}+1}{4k}\Big),
  \end{align*}
where $(a)$ follows from the fact that $(1+x)^{1/2} \le 1+ x/2$ for all $x\ge -1.$
This proves \eqref{eq:l1ub}, and the rest of the proposition follows immediately.
\end{IEEEproof}

\subsection{Comparison of our scheme with $k$-RR and $k$-RAPPOR}
In this section we compare our scheme to the two existing privatization schemes in the literature.
The $k$-RR scheme is the same as $\mathbi{Q}_{k,1}$ in this paper. The empirical estimator for $k$-RR scheme is given by \eqref{eq:emp} once we put $d=1.$ In the low-privacy regime, where $e^{\epsilon}\ge k,$ our choice of $d$ is $d=1,$ so in this regime our scheme coincides with the $k$-RR scheme.

To define the $k$-RAPPOR scheme \cite{Duchi13a, Erlingsson14}, let $\cY_{\RAP}=\{0,1\}^k.$
Given an input $i\in\cX,$ the output vector $Y$ is obtained by flipping each coordinate of $e_i$ independently with probability $1/(1+e^{\epsilon/2}),$ where $e_i$ is the $i$-th vector in the standard basis of $\mathbb{R}^k.$ Formally, the $k$-RAPPOR scheme 
$\mathbi{Q}_{\RAP}$ is defined as follows:
    $$
\mathbi{Q}_{\RAP}(y|i)= \Big( \frac{e^{\epsilon/2}y_i}{1+e^{\epsilon/2}} 
+\frac{1-y_i}{1+e^{\epsilon/2}}  \Big)
\prod_{j\neq i} \Big( \frac{e^{\epsilon/2}(1-y_j)}{1+e^{\epsilon/2}}
+\frac{y_j}{1+e^{\epsilon/2}} \Big)
   $$
for all $y=(y_1,y_2,\dots,y_k)\in \cY_{\RAP}$ and all $i\in\cX.$
The empirical estimator for the $k$-RAPPOR scheme is 
\begin{equation}\label{eq:empRAPPOR}
\hat{p}_i =\left( \frac{e^{\epsilon/2}+1}{e^{\epsilon/2}-1} \right) \frac{T_i}{n}-\frac{1}{e^{\epsilon/2}-1},
\end{equation}
where $T_i=\sum_{j=1}^n Y_i^{(j)}$ \cite{Duchi13a, Erlingsson14}. The associated risk values for the worst-case distribution $\mathbi{p}_U$ are given by
\begin{equation}\label{eq:RAPrisk}
\begin{aligned}
\underset{Y^n\sim \mathbi{m}_U^n}{\mathbb{E}} \ell_2^2(\hat{\mathbi{p}}(Y^n),\mathbi{p}_U)
&= \Big( 1 + \frac{k^2 e^{\epsilon/2}}{(k-1)(e^{\epsilon/2}-1)^2} \Big) \frac{k-1}{nk}, \\
\underset{Y^n\sim \mathbi{m}_U^n}{\mathbb{E}} \ell_1(\hat{\mathbi{p}}(Y^n),\mathbi{p}_U)
&=
\sqrt{\frac{2}{\pi n}\frac{(e^{\epsilon/2}+k-1)(e^{\epsilon/2}(k-1)+1)}{(e^{\epsilon/2}-1)^2} }
 + o\Big(\frac 1{\sqrt n}\Big),
\end{aligned}
\end{equation}
where $\mathbi{m}_U=\mathbi{p}_U\mathbi{Q}_{\RAP}$ \cite[Prop.~4]{Kairouz16}. In the high-privacy regime, where $\epsilon$ is close to $0,$ the $k$-RAPPOR scheme and its empirical estimator give order-optimal performance. More specifically, when $\epsilon$ is small and $k$ is large, the $\ell_2^2$ risk is approximately $\frac{4k}{n\epsilon^2},$ and the $\ell_1$ risk is approximately $\frac{2k}{\epsilon}\sqrt{\frac{2}{\pi n} }.$
At the same time, the authors of \cite{Duchi16} show that for $\epsilon$ close to $0$ the minimax risk \eqref{eq:rekn} behaves as   
   $$r_{\epsilon,k,n}^{\ell_2^2} = \Theta\Big(\frac{k}{n\epsilon^2}\Big) \text{ and }r_{\epsilon,k,n}^{\ell_1} =\Theta\Big(\frac{k}{\epsilon\sqrt{n}} \Big).
   $$
As a result, that the $k$-RAPPOR scheme gives order-optimal performance in high privacy regime.

To compare our scheme with $k$-RAPPOR in the high privacy regime,
let $\epsilon$ be small and $k$ be large. According to \eqref{eq:longub}-\eqref{eq:l1ub},
 the $\ell_2^2$ risk of our scheme is approximately $\frac{4k}{n\epsilon^2},$
and the $\ell_1$ risk is approximately $\frac{2k}{\epsilon}\sqrt{\frac{2}{\pi n} },$ which are exactly the same as those of $k$-RAPPOR scheme. Thus in the high privacy regime the proposed scheme does not improve over the known results.

\vspace*{.05in}
At the same time, the comparison is in favor of our schemes in the medium-privacy regime when $1\ll e^{\epsilon} \ll k.$ 
%Now let us compare our scheme with $k$-RR and $k$-RAPPOR in the medium privacy regime, where  We list the order of the $\ell_2^2$ risk and $\ell_1$ risk of the $3$ schemes for the worst case distribution $\mathbi{p}_U$ in the following table. 

\begin{proposition} The risks of the $k$-RR and $k$-RAPPOR schemes in the medium privacy regime are given in the following table.
\begin{center}
\begin{tabular}{| c | c | c |}
\hline
  & $\ell_2^2$ risk & $\ell_1$ risk \\ \hline
$\mathbi{Q}_{k, \lceil k/(e^{\epsilon}+1) \rceil}$ & $\Theta(\frac{k}{n e^{\epsilon}})$ & 
$\Theta(\frac{k}{\sqrt{n e^{\epsilon}}} ) $ \\[.05in] \hline 
$k$-RR & $\Theta (\frac{k^2}{n e^{2\epsilon}} \remove{\frac{k}{e^{\epsilon}}})$ &  $ \Theta (\frac{k^{3/2}}{\sqrt{n }e^{\epsilon}} \remove{\sqrt{\frac{k}{e^{\epsilon}}}} ) $\\ [.05in]\hline
$k$-RAPPOR & $\Theta(\frac{k}{n e^{\epsilon/2}} \remove{\sqrt{e^{\epsilon}}})$ & $\Theta(\frac{k}{\sqrt{n e^{\epsilon/2}}} 
\remove{e^{\epsilon/4}})$ \\[.05in]
\hline
\end{tabular}
\end{center}
\end{proposition}

We can make the claims of this proposition more specific by computing numerical bounds on the improvement of our scheme over the two existing schemes in the medium privacy regime.
We show that if $3.8 < \epsilon <\log(k/9) ,$ then the expected loss of our scheme is at most 
$50\%$ of the existing schemes under $\ell_2^2$ loss and at most $70\%$ of the existing schemes under $\ell_1$ loss.

To show this, let $r_{\RAP}^{\ell}(\mathbi{p})$ be the expected estimation loss of $k$-RAPPOR under its empirical estimator \eqref{eq:empRAPPOR} and let
$r_{\RR}^{\ell}(\mathbi{p})$ be the same for $k$-RR, both measured by loss function $\ell.$ 
Let  $r_{\OPT}^{\ell}(\mathbi{p})$ be
 the expected estimation loss under $\mathbi{Q}_{k,\lceil k/(e^{\epsilon}+1) \rceil}$ given in \eqref{eq:defQ} and its empirical estimator given in 
\eqref{eq:emp} for distribution $\mathbi{p}.$ 
(We omit parameters $n,k, \epsilon$ from the notation as they are clear from the context.)
We further define 
$$
r_{\RAP}^{\ell} = \max_{\mathbi{p}\in\Delta_k}  r_{\RAP}^{\ell}(\mathbi{p}), \quad
r_{\RR}^{\ell} = \max_{\mathbi{p}\in\Delta_k}  r_{\RR}^{\ell}(\mathbi{p}), \quad
r_{\OPT}^{\ell} = \max_{\mathbi{p}\in\Delta_k}  r_{\OPT}^{\ell}(\mathbi{p}).
$$
\begin{proposition}\label{prop:05}
If $\epsilon > 3.8$ and $k>9 e^{\epsilon},$ then
$$
r_{\OPT}^{\ell_2^2} <\frac{1}{2} \max( r_{\RAP}^{\ell_2^2} , r_{\RR}^{\ell_2^2}  ),
$$
and for large $n,$
$$
r_{\OPT}^{\ell_1} 
<0.7 \max( r_{\RAP}^{\ell_1} , r_{\RR}^{\ell_1}  ).
$$
\end{proposition}
The proof is given in Appendix \ref{ap:05}.
\begin{remark}
As discussed in \cite{Kairouz16}, along with the empirical estimator for the $k$-RR and $k$-RAPPOR schemes,
there are other estimators, for instance, the normalized estimator and the projected estimator. These estimators differ from the empirical estimator only when the latter gives some output which is not in $\Delta_k.$
Since the empirical estimator is unbiased, the probability of such events are exponentially small. As mentioned in the introduction, we are interested in the regime where $n$ is large, so the performance of different estimators only have exponentially small difference and can be neglected. This justifies our choice of only comparing the performance under empirical estimators.
\end{remark}

\section{Lower bound} \label{Sect:lb}
In this section, we give a tight lower bound on the minimax risk $r_{\epsilon,k,n}^{\ell}$ defined in
\eqref{eq:rekn}. 
Our argument consists of two steps. In the first step we establish that in order to obtain the optimal performance, we can restrict ourselves to the
privatization schemes with the so-called extremal configurations; cf. Theorem~\ref{thm:ext}. In this part we are motivated by a result in \cite{Kairouz14} which shows that a similar property holds for schemes optimal in terms of information theoretic utilities, such as mutual information between the input and the output. In the second step
we derive lower bounds on the risk that will establish order-optimality of the proposed privatization scheme. The main 
result of this section is given in the following theorem.
\begin{theorem} \label{Thm:lb}
If $n>\max(\frac{k^2(e^{\epsilon}+1)^2}{16(e^{\epsilon}-1)^2}, \frac{k^2}{2(e^{\epsilon}-1)}),$ then
\begin{align*}
r_{\epsilon, k,n}^{\ell_2^2}  \ge \frac{(k-1)(e^{\epsilon}+1)^2}{512n(e^{\epsilon}-1)^2}, \quad
r_{\epsilon, k,n}^{\ell_1}  \ge \frac{(k-1)(e^{\epsilon}+1)}{64\sqrt{n}(e^{\epsilon}-1)}
\text{~for~}  e^{\epsilon}<3,\\
r_{\epsilon, k,n}^{\ell_2^2}  \ge \frac{k-1}{64n(e^{\epsilon}-1)}, \quad
r_{\epsilon, k,n}^{\ell_1}  \ge \frac{k-1}{16\sqrt{2n(e^{\epsilon}-1)}}
\text{~for~} e^{\epsilon}\ge 3
\end{align*}
\end{theorem}

\subsection{Reduction to extremal configurations} We begin with showing that we only need to consider privatization schemes with finite output alphabet. The argument relies on the following technical lemma whose proof is given in Appendix~\ref{ap:approx}.

\begin{lemma}\label{lem:tech}
Let $P_1,P_2,\dots,P_k$ be probability measures defined on a measurable space $(\cY,\sigma(\cY)).$ For any partition of $\cY^n$ into a finite number of disjoint sets $\{B_i\}_{i=1}^N$ which are measurable with respect to the $n$-fold product $\sigma$-algebra 
$\sigma(\cY)^{\times n}:=\sigma(\cY)\times\sigma(\cY)\times\dots\times\sigma(\cY)$ and any $\alpha>0,$
there exists a partition of $\cY$ into a finite number of disjoint measurable sets $\{A_i\}_{i=1}^L\subseteq\sigma(\cY)$
and a partition of $\cY^n$ into disjoint sets $\{B'_i\}_{i=1}^N$ such that: 
\begin{enumerate}
%\item 
%$A_i\in \sigma(\cY)$ for every $i=1,2,\dots,L.$
\item The sets
$B'_i,i=1,2,\dots,N$ are measurable with respect to the $n$-fold finite product algebra 
$\sigma_F(\cY)^{\times n},$ where $\sigma_F(\cY)$ is the finite algebra generated by the sets $\{A_i\}_{i=1}^L;$
\item For any $i=1,2,\dots,N$ and any multi-index $\underline{j}= (j_1,j_2,\dots,j_n) \in [k]^n,$
  \begin{equation}\label{eq:alpha}
  | P_{\underline{j}} (B_i) - P_{\underline{j}}(B'_i ) | <\alpha,
  \end{equation}
where $P_{\underline{j}}:=P_{j_1}\times P_{j_2}\times \dots \times P_{j_n}$ is the $n$-fold product measure on the product measurable space $(\cY^n,\sigma(\cY)^{\times n}).$
\end{enumerate}
\end{lemma}

The next lemma establishes the fact that we do not need to look beyond finite output alphabets in our search for optimal schemes.
\begin{lemma} Let $\cD_{\epsilon,F}$ be the set of $\epsilon$-locally differentially private mechanisms with finite output alphabet.
For $\ell=\ell_2^2$ or $\ell_1,$
\begin{equation}\label{eq:redF}
r_{\epsilon,k,n}^{\ell} = \inf_{\mathbi{Q}\in\cD_{\epsilon,F}}r_{k,n}^{\ell} (\mathbi{Q}),
\end{equation}
\end{lemma}
\begin{IEEEproof}
Define a clipping function $g:\mathbb{R} \to [0,1]$ as follows: 
  $$
  g(x)=\begin{cases} 0 &x<0,\\x &0\le x< 1,\\1 &x\ge 1,\end{cases}
  $$
and define its extension $g_k: \mathbb{R}^k \to [0,1]^k$ as $g_k((v_1,v_2,\dots,v_k))=(g(v_1),g(v_2),\dots,g(v_k))$ for all $(v_1,\dots,v_k)\in \mathbb{R}^k.$
It is clear that $\ell(\hat{\mathbi{p}}(Y^n), \mathbi{p}) \ge \ell(g_k(\hat{\mathbi{p}}(Y^n)), \mathbi{p})$ for both $\ell=\ell_1$ and $
\ell=\ell_2^2.$ This implies that the optimal estimator should take values in $[0,1]^k$ instead of $\mathbb{R}^k.$ Thus, in the proof 
below we only need to consider estimators taking values in $[0,1]^k.$

It suffices to show that for any $\alpha>0,$ any $\mathbi{Q}\in\cD_{\epsilon}$ with some output alphabet $\cY,$ and any estimator $\hat{\mathbi{p}}:\cY^n\to [0,1]^k,$ we can find a private mechanism
$\mathbi{Q}_F\in\cD_{\epsilon,F}$ with some finite output alphabet $\cY_F,$ and an estimator $\hat{\mathbi{p}}_F:\cY_F^n\to [0,1]^k,$ such that
\begin{equation}\label{eq:obj}
\underset{Y_F^n\sim \mathbi{m}_F^n}{\mathbb{E}} \ell(\hat{\mathbi{p}}_F(Y_F^n), \mathbi{p})
\le 
\underset{Y^n\sim \mathbi{m}^n}{\mathbb{E}} \ell(\hat{\mathbi{p}}(Y^n), \mathbi{p}) +\alpha
\text{~for all~} \mathbi{p}\in \Delta_k,
\end{equation}
where $\mathbi{m}_F = \mathbi{p} \mathbi{Q}_F,$ and $\mathbi{m} = \mathbi{p} \mathbi{Q}.$

Given an integer $t,$ we partition the interval $[0,1]$ into $t$ disjoint sets
$\{C_i\}_{i=1}^t,$ where
$$
C_i = [(i-1)/t, i/t) \text{~for all~} i=1,2,\dots,t-1, \text{~and~}
C_{t} = [(t-1)/t,1].
$$
We partition $[0,1]^k$ into $t^k$ disjoint sets 
$\{C_{{u}_1}\times C_{{u}_2}\times \dots \times C_{{u}_k}: 1\le {u}_1,{u}_2,\dots,{u}_k \le t \}.$ Define
the multi-index $\underline{u}=(u_1,u_2,\dots,u_k)\in[t]^k$ and the set
$B_{\underline{u}}:=\hat{\mathbi{p}}^{-1}( C_{u_1}\times C_{u_2}\times \dots \times C_{u_k} )$.
 Clearly, the collection $\{B_{{\underline{u}}},{\underline{u}}\in{[t]^k}\}$ forms a partition of $\cY^n.$
As a result, 
\begin{align*}
\underset{Y^n\sim \mathbi{m}^n}{\mathbb{E}} \ell(\hat{\mathbi{p}}(Y^n), \mathbi{p})
& =\sum_{x^n\in\cX^n} \left( \mathbi{p}^n(x^n) \int_{y^n\in\cY^n} 
\ell(\hat{\mathbi{p}}(y^n),\mathbi{p}) d \mathbi{Q}^n(y^n|x^n) \right) \\
& = \sum_{x^n\in\cX^n} \Big( \mathbi{p}^n(x^n)  \sum_{{\underline{u}}\in{[t]^k}}\int_{y^n\in B_{{\underline{u}}}} 
\ell(\hat{\mathbi{p}}(y^n),\mathbi{p}) d\mathbi{Q}^n(y^n|x^n) \Big),
\end{align*}
where the integrals are computed with respect to the product measure $\mathbi{Q}^n(\cdot |x^n).$
%and the variable of integration is $y^n.$
Consequently, for both $\ell=\ell_1$ and $\ell_2^2$ we have
\begin{equation}\label{eq:diffp}
\begin{aligned}
  \Big| \underset{Y^n\sim \mathbi{m}^n}{\mathbb{E}} \ell(\hat{\mathbi{p}}(Y^n), \mathbi{p})
- &\sum_{x^n\in\cX^n} \Big( \mathbi{p}^n(x^n) \sum_{{\underline{u}}\in{[t]^k}}  \ell({\underline{u}} /t, \mathbi{p}) \mathbi{Q}^n(B_{{\underline{u}}}|x^n) \Big) \Big| \\
\le & \sum_{x^n\in\cX^n} \Big( \mathbi{p}^n(x^n)  \sum_{{\underline{u}}\in{[t]^k}}\int_{y^n\in B_{{\underline{u}}}} 
\Big|\ell(\hat{\mathbi{p}}(y^n),\mathbi{p})-  \ell({\underline{u}} /t, \mathbi{p})\Big| d\mathbi{Q}^n(y^n|x^n) \Big) \\
\le &\max_{{\underline{u}}\in{[t]^k}} \Big( \sup_{y^n \in B_{{\underline{u}}}} \Big|\ell(\hat{\mathbi{p}}(y^n),\mathbi{p})-  \ell({\underline{u}} /t, \mathbi{p})\Big| \Big) \overset{(a)}{\le} \frac{2k}{t}
\end{aligned}
\end{equation}
Note that apart from inequality $(a),$ all the other inequalities above do not depend on the choice of $\ell.$ 
The inequality $(a)$ depends on the choice of $\ell=\ell_1,\ell_2^2,$ and is obtained by simple calculation.

Notice that $\mathbi{Q}(\cdot | 1), \mathbi{Q}(\cdot | 2), \dots, \mathbi{Q}(\cdot | k)$ are $k$ probability measures on the same measurable space $(\cY,\sigma(\cY)),$
and $\mathbi{Q}^n (\cdot | x^n)=\mathbi{Q}(\cdot |x^{(1)}) \times \mathbi{Q}(\cdot |x^{(2)}) \times \dots \times \mathbi{Q}(\cdot |x^{(n)})$ is the $n$-fold product measure on the $n$-fold product measurable space $(\cY^n, \sigma(\cY)\times \sigma(\cY) \times \dots \times \sigma(\cY)).$
According to Lemma~\ref{lem:tech}, for any $\alpha'>0,$ we can find a partition of $\cY$ into a finite number of disjoint sets $\{A_i\}_{i=1}^L$ together with a partition of $\cY^n$ into disjoint sets $\{B'_{{\underline{u}}}\}_{{\underline{u}}\in{[t]^k}}$
such that 
\begin{enumerate}
\item $A_i\in \sigma(\cY)$ for every $i=1,2,\dots,L.$
\item $B'_{{\underline{u}}}$ are measurable with respect to the $n$-fold product $\sigma$-algebra $\sigma_F(\cY)\times \sigma_F(\cY)\times\dots\times\sigma_F(\cY)$ for every ${\underline{u}}\in{[t]^k},$ where $\sigma_F(\cY)$ is the finite 
$\sigma$-algebra generated by $\{A_i\}_{i=1}^L.$
\item For every ${\underline{u}}\in{[t]^k}$ and every $x^n\in\cX^n,$
\begin{equation}\label{eq:prime}
| \mathbi{Q}^n(B_{{\underline{u}}} |x^n) - \mathbi{Q}^n(B'_{{\underline{u}}} |x^n) |<\alpha'.
\end{equation}
\end{enumerate}

By definition, $\sigma_F(\cY)\times \sigma_F(\cY)\times\dots\times\sigma_F(\cY)$ is generated by the following finite partition of $\cY^n:$
$$
\{A_{\nu_1}\times A_{\nu_2}\times \dots\times A_{\nu_n}: 1\le \nu_1,\nu_2,\dots,\nu_n\le L\}.
$$
For every $\underline{\nu}=(\nu_1,\nu_2,\dots,\nu_n) \in [L]^n,$ there is a unique ${\underline{u}}\in{[t]^k}$ such that 
$A_{\nu_1}\times A_{\nu_2}\times \dots\times A_{\nu_n}\subseteq B'_{{\underline{u}}}.$
Define a function $f:{[L]^n}\to{[t]^k}$ as follows: $f(\underline{\nu}), \underline{\nu}\in{[L]^n}$ is the unique vector in ${[t]^k}$ such that
$A_{\nu_1}\times A_{\nu_2}\times \dots\times A_{\nu_n}\subseteq B'_{f(\nu)}.$ 

Further, define $\mathbi{Q}_F:\cX\to \cY_F=\{1,2,\dots,L\}$ and 
$\hat{\mathbi{p}}_F:\cY_F^n\to [0,1]^k$ as follows:
\begin{align*}
\mathbi{Q}_F(i|x)=\mathbi{Q}(A_i|x) &\text{~for all~} i\in \cY_F \text{~and~} x\in\cX,\\
\hat{\mathbi{p}}_F(\underline{\nu})=\frac{1}{t} f(\underline{\nu}) &\text{~for all~} \underline{\nu}\in \cY_F^n.
\end{align*}
It is clear that $\mathbi{Q}_F\in\cD_{\epsilon,F}.$
Also note that
$
\mathbi{Q}_F^n(\underline{\nu}|x^n)=\mathbi{Q}^n(A_{\nu_1}\times A_{\nu_2}\times \dots\times A_{\nu_n}|x^n)
$ for all $\underline{\nu}\in{[L]^n}$ and $x^n\in\cX^n.$
Therefore,
$$
\mathbi{Q}_F^n(f^{-1}({\underline{u}})|x^n)= \mathbi{Q}^n(B'_{{\underline{u}}}|x^n) \text{~for all~} {\underline{u}}\in{[t]^k} \text{~and~} x^n\in\cX^n.
$$
Thus we have
\begin{align*}
\underset{Y_F^n\sim \mathbi{m}_F^n}{\mathbb{E}} \ell(\hat{\mathbi{p}}_F(Y_F^n), \mathbi{p})
&= \sum_{x^n\in\cX^n} \left( \mathbi{p}^n(x^n) \sum_{\underline{\nu}\in{[L]^n}}  \ell(\hat{\mathbi{p}}_F(\underline{\nu}), \mathbi{p})  \mathbi{Q}_F^n(\underline{\nu}|x_n)  \right) \\
&= \sum_{x^n\in\cX^n} \left( \mathbi{p}^n(x^n) \sum_{{\underline{u}}\in{[t]^k}}  \ell({\underline{u}} /t, \mathbi{p})  \mathbi{Q}_F^n(f^{-1}({\underline{u}})|x_n) \right) \\
&= \sum_{x^n\in\cX^n} \left( \mathbi{p}^n(x^n) \sum_{{\underline{u}}\in{[t]^k}}  \ell({\underline{u}} /t, \mathbi{p}) \mathbi{Q}^n(B'_{{\underline{u}}}|x^n) \right).
\end{align*}
For both $\ell=\ell_1$ and $\ell_2^2,$ we have that $\ell(v,v') \le k$ for all $v,v'\in[0,1]^k.$ Thus for 
$\ell=\ell_1$ or $\ell_2^2,$ we have the following inequality
\begin{equation}\label{eq:diffpF}
\begin{aligned}
 \Big| \underset{Y_F^n\sim \mathbi{m}_F^n}{\mathbb{E}} \ell(\hat{\mathbi{p}}_F(Y_F^n), \mathbi{p})
&-  \sum_{x^n\in\cX^n} \Big( \mathbi{p}^n(x^n) \sum_{{\underline{u}}\in{[t]^k}}  \ell({\underline{u}} /t, \mathbi{p}) \mathbi{Q}^n(B_{{\underline{u}}}|x^n) \Big) \Big|  \\
\le & \sum_{x^n\in\cX^n} \Big( \mathbi{p}^n(x^n) \sum_{{\underline{u}}\in{[t]^k}}  \ell({\underline{u}} /t, \mathbi{p}) \Big| \mathbi{Q}^n(B_{{\underline{u}}}|x^n) - \mathbi{Q}^n(B'_{{\underline{u}}}|x^n) \Big| \Big) \\
\overset{(a)}{\le} & \sum_{x^n\in\cX^n} \Big( \mathbi{p}^n(x^n) \sum_{{\underline{u}}\in{[t]^k}}  \ell({\underline{u}} /t, \mathbi{p}) \alpha' \Big) \\
\le & t^k k\alpha' \sum_{x^n\in\cX^n}  \mathbi{p}^n(x^n) = t^k k \alpha',
\end{aligned}
\end{equation}
where $(a)$ follows from \eqref{eq:prime}.
Using inequalities \eqref{eq:diffp} and \eqref{eq:diffpF} together with the triangle inequality, we deduce that
   $$
\Big| \underset{Y_F^n\sim \mathbi{m}_F^n}{\mathbb{E}} \ell(\hat{\mathbi{p}}_F(Y_F^n), \mathbi{p})
- 
\underset{Y^n\sim \mathbi{m}^n}{\mathbb{E}} \ell(\hat{\mathbi{p}}(Y^n), \mathbi{p})  \Big|
\le \frac{2k}{t} + t^k k\alpha'.
   $$
By setting $t> (4k)/\alpha$ and $\alpha'<\alpha/(2t^k k),$ we obtain the desired result \eqref{eq:obj} and thus complete the proof of the lemma.
\end{IEEEproof}

We continue to implement the plan laid out in the beginning of the section. The next step is to show that we can further restrict ourselves to the following set of private schemes with extremal configurations:
$$
\cD_{\epsilon,E}=\biggl\{ \mathbi{Q}\in\cD_{\epsilon,F}: 
\frac{\mathbi{Q}(y|x)}{\min_{x'\in\cX}\mathbi{Q}(y|x') } \in \{1,e^{\epsilon}\}
\text{~for all~} x\in\cX \text{~and all~} y\in\cY \biggr\}.
$$ 

Before we show that we only need to consider $\mathbi{Q}\in\cD_{\epsilon,E},$ we establish the following easy claim.
 \begin{lemma}\label{lemcv} Let $\cA:=[1,e^{\epsilon}]^k$ and $\cB:=\{1,e^{\epsilon}\}^k.$
Every vector in $\cA$ can be written as a convex combination of vectors in $\cB.$ 
\end{lemma}
\begin{IEEEproof} Basically this lemma says that every point in the cube is a convex combination of its $2^k$ vertices, which is
of course obvious. To prove this formally, define a function $f:\cA \to \{0,1,2,\dots,k\}$ as 
$f(v)=|\{i\in\{1,2,\dots,k\}:v_i \neq 1\text{~or~} e^{\epsilon}\}|$
for all vectors $v=(v_1,v_2,\dots,v_k) \in \cA.$
We prove the claim by induction on $f(v).$
Clearly, if $f(v)=0,$ then $v\in\cB.$ This establishes the induction basis. Now suppose that the claim holds true for every vector $v'$ such that $f(v') = i-1$ and let $v$ be such that $f(v)=i.$ Without loss of generality, suppose that $v_1\neq 1$ or $e^{\epsilon}.$ Then we can write 
\begin{equation}\label{eq:expv}
v=\frac{e^{\epsilon}-v_1}{e^{\epsilon}-1}v' + \frac{v_1 - 1}{e^{\epsilon}-1} v'',
\end{equation}
where
$v'=(1,v_2,v_3,\dots,v_k)$ and $v''=(e^{\epsilon},v_2,v_3,\dots,v_k).$ Since $f(v')=f(v'')=i-1,$ by induction hypothesis, we can write both $v'$ and $v''$ as convex combinations of vectors in $\cB.$ Substituting these expressions for $v'$ and $v''$ into \eqref{eq:expv}, we can write $v$ as a convex combination of vectors in $\cB.$ This proves the induction step.
\end{IEEEproof}

\begin{theorem}\label{thm:ext}
For $\ell=\ell_2^2$ and $\ell_1,$
\begin{equation}\label{eq:red}
r_{\epsilon,k,n}^{\ell} = \inf_{\mathbi{Q}\in\cD_{\epsilon,E}} r_{k,n}^{\ell} (\mathbi{Q}).
\end{equation}
\end{theorem}
\begin{IEEEproof} We already know from \eqref{eq:redF} that finite $\cY$ suffices. To prove the lemma we only need to show that for any 
$\mathbi{Q}\in\cD_{\epsilon, F}$ with some finite output alphabet $\cY,$ and any estimator 
$\hat{\mathbi{p}}:\cY^n\to \mathbb{R}^k,$ we can find a private mechanism
$\mathbi{Q}_E\in\cD_{\epsilon,E}$ with some finite output alphabet $\cY_E,$ and an estimator $\hat{\mathbi{p}}_E:\cY_E^n\to \mathbb{R}^k,$ such that
\begin{equation}\label{eq:objE}
\underset{Y_E^n\sim \mathbi{m}_E^n}{\mathbb{E}} \ell(\hat{\mathbi{p}}_E(Y_E^n), \mathbi{p}) =
\underset{Y^n\sim \mathbi{m}^n}{\mathbb{E}} \ell(\hat{\mathbi{p}}(Y^n), \mathbi{p}) 
\text{~for all~} \mathbi{p}\in \Delta_k,
\end{equation}
where $\mathbi{m}_E = \mathbi{p} \mathbi{Q}_E$ and $\mathbi{m} = \mathbi{p} \mathbi{Q}.$

Without loss of generality, suppose that $\cY=\{0,1,\dots,L-1\}$ for some integer $L.$ For $j\in\cY,$ let $Q_j=\min_{x\in\cX} \mathbi{Q}(j|x).$ Since $\mathbi{Q}$ is $\epsilon$-locally differentially private, the vector
  $$
  \frac 1{Q_j}(\mathbi{Q}(j|1),\mathbi{Q}(j|2),\dots,\mathbi{Q}(j|k)) \in \cA
  $$
(recall that $\cA=[1,e^\epsilon]$, Lemma~\ref{lemcv}).   According to Lemma~\ref{lemcv},
we can write this vector as 
  $$\frac 1{Q_j}(\mathbi{Q}(j|1),\mathbi{Q}(j|2),\dots,\mathbi{Q}(j|k))=\sum_{i=0}^{2^k-1}w_{j,i}\mathbi{b}_i,
  $$ where 
$\{w_{j,i}\}_{i=0}^{2^k-1}$ are nonnegative coefficients that add to one, and
$\mathbi{b}_0,\mathbi{b}_1,\dots,\mathbi{b}_{2^k-1}$ are the $2^k$ vectors in the cube $\cB$ (labeled in arbitrary order).

Now define $\mathbi{Q}_E:\cX \to \cY_E=\{0,1,\dots,2^k L-1\}$ as follows:
$$
(\mathbi{Q}_E(2^k j+i|1), \mathbi{Q}_E(2^k j+i|2), \dots, \mathbi{Q}_E(2^k j+i|k))
= Q_j w_{j,i}\mathbi{b}_i \text{~for all~} j\in\cY \text{~and~} i=0,1,\dots,2^k-1.
$$
Clearly $\mathbi{Q}_E$ is a valid conditional distribution. We define a function $f:\cY_E\to\cY$ as 
$f(y_E)=\lfloor y_E/ 2^k \rfloor$ for all $y_E\in \cY_E.$  It is easy to check that $f(Y_E)$ has distribution $\mathbi{p} \mathbi{Q}.$ In other words, we can use the output of $\mathbi{Q}_E$ to reproduce the output of $\mathbi{Q}$ with exactly the same distribution.
 Given an estimator $\hat{\mathbi{p}}:\cY^n\to \mathbb{R}^k,$ we define $\hat{\mathbi{p}}_E:\cY_E^n\to \mathbb{R}^k$ as
$\hat{\mathbi{p}}_E(y_E^n)= \hat{\mathbi{p}}((f(y_E^{(1)}), f(y_E^{(2)}), \dots, f(y_E^{(n)}) ))$
for all $y_E^n=(y_E^{(1)}, y_E^{(2)}, \dots, y_E^{(n)}) \in\cY_E^n.$ The pair $(\mathbi{Q}_E,\hat{\mathbi{p}}_E)$ satisfies \eqref{eq:objE}. This completes the proof.
\end{IEEEproof}

\subsection{Derivation of the lower bound: Proof of Theorem \ref{Thm:lb}} 
In the previous subsection we have prepared ground for the proof of the lower bounds on $r_{\epsilon, k,n}^{\ell}$ stated
in Theorem \ref{Thm:lb}.
In the classical (non-private) minimax estimation problem, one standard approach to the proof of lower bounds on the
minimax risk of estimation is Assouad's method \cite{Assouad83} (see also \cite{tsybakov09}).
Duchi et al. \cite{Duchi16} developed Assouad's method in the private setting. 
In our proof we refine the technique in \cite{Duchi16} to obtain a tight lower bound in the regime $e^{\epsilon} \ll k.$ The first
steps in the proof are inspired by the approach in \cite{Duchi16}.
%
%\begin{theorem} \label{Thm:lb}
%If $n>\max(\frac{k^2(e^{\epsilon}+1)^2}{16(e^{\epsilon}-1)^2}, \frac{k^2}{2(e^{\epsilon}-1)}),$ then
%\begin{align*}
%r_{\epsilon, k,n}^{\ell_2^2}  \ge \frac{(k-1)(e^{\epsilon}+1)^2}{512n(e^{\epsilon}-1)^2}, \quad
%r_{\epsilon, k,n}^{\ell_1}  \ge \frac{(k-1)(e^{\epsilon}+1)}{64\sqrt{n}(e^{\epsilon}-1)}
%\text{~for~}  e^{\epsilon}<3,\\
%r_{\epsilon, k,n}^{\ell_2^2}  \ge \frac{k-1}{64n(e^{\epsilon}-1)}, \quad
%r_{\epsilon, k,n}^{\ell_1}  \ge \frac{k-1}{16\sqrt{2n(e^{\epsilon}-1)}}
%\text{~for~} e^{\epsilon}\ge 3
%\end{align*}
%\end{theorem}
%\begin{IEEEproof}

Let $\delta\in[0,1].$ We begin with the case of even $k$ (the proof for $k$ odd requires only a minor modification).
Let $\cV=\{-1,1\}^{k/2},$ and for $\nu = (\nu_1,\dots,\nu_{k/2}) \in\cV$ let $\mathbi{p}_{\nu}$ be the distribution
$$
\mathbi{p}_{\nu} := \mathbi{p}_U + \frac{\delta}{k}
\left[\begin{array}{@{}c@{}}\nu \\ -\nu \end{array}\right]
\in\Delta_k.
$$
For any privatization mechanism $\mathbi{Q}:\cX\to\cY$ and any estimator $\hat{\mathbi{p}}:\cY^n\to \mathbb{R}^k,$
$$
\sup_{\mathbi{p}\in \Delta_k} 
\underset{Y^n\sim (\mathbi{p}\mathbi{Q})^n}{\mathbb{E}} \ell(\hat{\mathbi{p}}(Y^n), \mathbi{p})
\ge \sup_{\nu\in\cV}
\underset{Y^n\sim (\mathbi{p}_{\nu}\mathbi{Q})^n}{\mathbb{E}} \ell(\hat{\mathbi{p}}(Y^n), \mathbi{p}_{\nu})
\ge \frac{1}{|\cV|}\sum_{\nu\in\cV} \underset{Y^n\sim (\mathbi{p}_{\nu}\mathbi{Q})^n}{\mathbb{E}} \ell(\hat{\mathbi{p}}(Y^n), \mathbi{p}_{\nu}).
$$
Consequently,
\begin{equation}\label{eq:rQ}
r_{k,n}^{\ell} (\mathbi{Q}) \ge \inf_{\hat{\mathbi{p}}} 
\frac{1}{|\cV|}\sum_{\nu\in\cV} \underset{Y^n\sim (\mathbi{p}_{\nu}\mathbi{Q})^n}{\mathbb{E}} \ell(\hat{\mathbi{p}}(Y^n), \mathbi{p}_{\nu}).
\end{equation}
According to \eqref{eq:red}, we only need to prove that the lower bounds on the risk hold for all $\mathbi{Q}\in \cD_{\epsilon,E}.$ 

\subsubsection{Loss function $\ell_2^2$}
We begin with the case of the loss function $\ell=\ell_2^2.$ Below we use the notation $\hat{\mathbi{p}}(y^n):=(\hat{p}_1(y^n), \hat{p}_2(y^n), \dots, \hat{p}_k(y^n)).$
For every estimator $\hat{\mathbi{p}}$ and every $y^n\in\cY^n,$ we have
  \begin{equation}\label{eq:initmul}
\ell_2^2(\hat{\mathbi{p}}(y^n), \mathbi{p}_{\nu})
\ge \sum_{j=1}^{k/2} \Big(\hat{p}_j(y^n)-\Big(\frac{1}{k} + \frac{\delta \nu_j}{k}\Big)\Big)^2
\ge \frac{\delta^2}{k^2} \sum_{j=1}^{k/2} \mathbbm{1}\Big\{\sign\Big(\hat{p}_j(y^n)-\frac{1}{k}\Big) \neq \nu_j\Big\},
   \end{equation}
where $\sign(x)=1$ for all $x\ge 0,$ and $\sign(x)=-1$ for all $x < 0.$
For $j=1,2,\dots,k/2,$
define the functions $g_j:\cY^n\to\{-1,1\}$ as $g_j(y^n)=\sign(\hat{p}_j(y^n)-\frac{1}{k})$ for all $y^n\in\cY^n.$
(Note that the function $g_j$ depends on the estimator $\hat{\mathbi{p}}.$ We will omit this dependence from the notation for simplicity.)
For $j=1,2\dots,k/2,$ define the mixture distributions\footnote{In \cite{Duchi16}, the authors treat $\mathbi{m}_{+j}^M$ and $\mathbi{m}_{-j}^M$ as product distributions, which is obviously not the case. This mistake enables them to claim better constants in their lower bound than in ours.}
\begin{equation}\label{eq:defmix}
\mathbi{m}_{+j}^M=\frac{2}{|\cV|}\sum_{\nu:\nu_j=1} (\mathbi{p}_{\nu}\mathbi{Q})^n, \quad
\mathbi{m}_{-j}^M=\frac{2}{|\cV|}\sum_{\nu:\nu_j=-1} (\mathbi{p}_{\nu}\mathbi{Q})^n.
\end{equation}

Then for every estimator $\hat{\mathbi{p}},$
\begin{equation}\label{eq:bdep}
\begin{aligned}
\frac{1}{|\cV|}\sum_{\nu\in\cV} &\underset{Y^n\sim (\mathbi{p}_{\nu}\mathbi{Q})^n}{\mathbb{E}} \ell_2^2(\hat{\mathbi{p}}(Y^n), \mathbi{p}_{\nu})
\ge \frac{\delta^2}{k^2} \frac{1}{|\cV|}\sum_{\nu\in\cV} \underset{Y^n\sim (\mathbi{p}_{\nu}\mathbi{Q})^n}{\mathbb{E}}
 \sum_{j=1}^{k/2} \mathbbm{1}\{g_j(Y^n) \neq \nu_j\} \\
= &\frac{\delta^2}{k^2} 
\sum_{j=1}^{k/2} \frac{1}{|\cV|} \sum_{\nu\in\cV}
 \underset{Y^n\sim (\mathbi{p}_{\nu}\mathbi{Q})^n}{\mathbb{E}} \mathbbm{1}\{g_j(Y^n) \neq \nu_j\} \\
= &\frac{\delta^2}{k^2} 
\sum_{j=1}^{k/2}  \Big( \frac{1}{|\cV|} \sum_{\nu:\nu_j=1}
 \underset{Y^n\sim (\mathbi{p}_{\nu}\mathbi{Q})^n}{\mathbb{E}} \mathbbm{1}\{g_j(Y^n) = -1 \}
+ \frac{1}{|\cV|} \sum_{\nu:\nu_j=-1} 
 \underset{Y^n\sim (\mathbi{p}_{\nu}\mathbi{Q})^n}{\mathbb{E}} \mathbbm{1}\{g_j(Y^n) = 1\} \Big) \\
= &\frac{\delta^2}{k^2} 
\sum_{j=1}^{k/2} \frac{1}{2}  \Big( 
 \underset{Y^n\sim \mathbi{m}_{+j}^M}{\mathbb{E}} \mathbbm{1}\{g_j(Y^n) = -1 \} +
 \underset{Y^n\sim \mathbi{m}_{-j}^M}{\mathbb{E}} \mathbbm{1}\{g_j(Y^n) = 1\} \Big) \\
\ge  &\frac{\delta^2}{k^2} 
\sum_{j=1}^{k/2} \frac{1}{2} \inf_{\psi} \Big( 
 \underset{Y^n\sim \mathbi{m}_{+j}^M}{\mathbb{E}} \mathbbm{1}\{\psi(Y^n) = -1 \} +
 \underset{Y^n\sim \mathbi{m}_{-j}^M}{\mathbb{E}} \mathbbm{1}\{\psi(Y^n) = 1\} \Big) \\
= &\frac{\delta^2}{2k^2} 
\sum_{j=1}^{k/2} \inf_{\psi} \left( 
 \mathbi{m}_{+j}^M (\psi(Y^n) = -1 ) +
 \mathbi{m}_{-j}^M (\psi(Y^n) = 1) \right),
\end{aligned}
\end{equation}
where the infimum above is taken over all the functions mapping from $\cY^n$ to $\{1,-1\}.$
Define the set $A_{\psi}=\{y^n\in \cY^n: \psi(y^n)=-1\}.$ Then 
$A_{\psi}^c=\{y^n\in \cY^n: \psi(y^n)=1\}.$ We have
\begin{equation}\label{eq:totalv}
\begin{aligned}
\inf_{\psi} \big( 
 \mathbi{m}_{+j}^M (\psi(Y^n) = -1 ) &+
 \mathbi{m}_{-j}^M (\psi(Y^n) = 1) \big)\\
&= \inf_{\psi} \left( 
 \mathbi{m}_{+j}^M (A_{\psi}) +
 \mathbi{m}_{-j}^M (A_{\psi}^c ) \right)
=  \inf_{\psi} \left( 1 -
 \left( \mathbi{m}_{-j}^M (A_{\psi}) 
-  \mathbi{m}_{+j}^M (A_{\psi}) \right) \right) \\
= &\inf_{A\subseteq \cY^n} \left( 1 -
 \left( \mathbi{m}_{-j}^M (A) 
-  \mathbi{m}_{+j}^M (A) \right) \right)
= 1 -
\sup_{A\subseteq \cY^n} \left( \mathbi{m}_{-j}^M (A) 
-  \mathbi{m}_{+j}^M (A) \right) \\
= & 1- \|\mathbi{m}_{-j}^M - \mathbi{m}_{+j}^M \|_{\TV}.
\end{aligned}
\end{equation}

Let $\{e_j\}_{j=1}^{k/2}$ be a standard basis of $\mathbb{R}^{k/2}.$ By definition \eqref{eq:defmix}, we have
    \begin{align}
\|\mathbi{m}_{-j}^M - \mathbi{m}_{+j}^M \|_{\TV}
&\le \frac{2}{|\cV|}\sum_{\nu:\nu_j=-1} \|(\mathbi{p}_{\nu}\mathbi{Q})^n
- (\mathbi{p}_{\nu+2e_j}\mathbi{Q})^n  \|_{\TV}\nonumber\\
&\le \sup_{\nu:\nu_j=-1} \|(\mathbi{p}_{\nu}\mathbi{Q})^n
- (\mathbi{p}_{\nu+2e_j}\mathbi{Q})^n  \|_{\TV}\label{eq:bdtv}
    \end{align}

Combining \eqref{eq:bdep}-\eqref{eq:bdtv}, we obtain
\begin{equation}\label{eq:ETV}
\frac{1}{|\cV|}\sum_{\nu\in\cV} \underset{Y^n\sim (\mathbi{p}_{\nu}\mathbi{Q})^n}{\mathbb{E}} \ell_2^2(\hat{\mathbi{p}}(Y^n), \mathbi{p}_{\nu})
\ge \frac{\delta^2}{2k^2} 
\sum_{j=1}^{k/2} \Big( 1- \sup_{\nu:\nu_j=-1} \|(\mathbi{p}_{\nu}\mathbi{Q})^n
- (\mathbi{p}_{\nu+2e_j}\mathbi{Q})^n  \|_{\TV} \Big).
\end{equation}
We also have the following inequality,
  \begin{align}
\sum_{j=1}^{k/2}  \sup_{\nu:\nu_j=-1} \|(\mathbi{p}_{\nu}\mathbi{Q})^n
- (\mathbi{p}_{\nu+2e_j}\mathbi{Q})^n  \|_{\TV} 
&
\overset{(a)}{\le}  \sqrt{\frac{k}{2} \Big(\sum_{j=1}^{k/2} 
\sup_{\nu:\nu_j=-1} \|(\mathbi{p}_{\nu}\mathbi{Q})^n
- (\mathbi{p}_{\nu+2e_j}\mathbi{Q})^n  \|_{\TV}^2 \Big)} \nonumber\\
&\overset{(b)}{\le}  \sqrt{\frac{k}{4} \Big(\sum_{j=1}^{k/2} 
\sup_{\nu:\nu_j=-1} D_{\kl} \big( (\mathbi{p}_{\nu}\mathbi{Q})^n||
(\mathbi{p}_{\nu+2e_j}\mathbi{Q})^n \big) \Big)}  \nonumber\\
&\le \sqrt{\frac{kn}{4} \Big(\sum_{j=1}^{k/2} 
\sup_{\nu:\nu_j=-1} D_{\kl} \big( \mathbi{p}_{\nu}\mathbi{Q} ||
\mathbi{p}_{\nu+2e_j}\mathbi{Q} \big) \Big)}, \label{eq:CSP}
   \end{align}
where $(a)$ follows from Cauchy-Schwarz inequality, and $(b)$ follows from Pinsker's inequality. 
Substituting \eqref{eq:CSP} into \eqref{eq:ETV}, we deduce that for every estimator $\hat{\mathbi{p}},$
   $$
\frac{1}{|\cV|}\sum_{\nu\in\cV} \underset{Y^n\sim (\mathbi{p}_{\nu}\mathbi{Q})^n}{\mathbb{E}} \ell_2^2(\hat{\mathbi{p}}(Y^n), \mathbi{p}_{\nu})
\ge \frac{\delta^2}{4k} \Biggl(1- \sqrt{\frac{n}{k} \Big(\sum_{j=1}^{k/2} 
\sup_{\nu:\nu_j=-1} D_{\kl} \Big( \mathbi{p}_{\nu}\mathbi{Q} ||
\mathbi{p}_{\nu+2e_j}\mathbi{Q} \Big) \Big)} \Biggr).
   $$
Going back to \eqref{eq:rQ}, we now obtain the bound
\begin{equation}\label{eq:bdrq}
r_{k,n}^{\ell_2^2} (\mathbi{Q}) \ge
\frac{\delta^2}{4k} \biggl(1- \sqrt{\frac{n}{k} \Big(\sum_{j=1}^{k/2} 
\sup_{\nu:\nu_j=-1} D_{\kl} \Big( \mathbi{p}_{\nu}\mathbi{Q} ||
\mathbi{p}_{\nu+2e_j}\mathbi{Q} \Big) \Big)} \biggr).
\end{equation}
Let $\mathbi{p}_{\nu}(i),i=1,2,\dots,k$ be the $i$-th coordinate of $\mathbi{p}_{\nu}.$ 
For $\mathbi{Q}\in \cD_{\epsilon,E},$
\begin{align}
\sum_{j=1}^{k/2} 
\max_{\nu:\nu_j=-1} &D_{\kl} \big( \mathbi{p}_{\nu}\mathbi{Q} ||
\mathbi{p}_{\nu+2e_j}\mathbi{Q}\big) \nonumber \\
&=  \sum_{j=1}^{k/2} 
\max_{\nu:\nu_j=-1} \sum_{y\in\cY} \Big( \Big(\sum_{i=1}^k \mathbi{p}_{\nu}(i)\mathbi{Q}(y|i) \Big)
\log \frac{\sum_{i=1}^k \mathbi{p}_{\nu}(i)\mathbi{Q}(y|i)}
{\sum_{i=1}^k \mathbi{p}_{\nu+2e_j}(i)\mathbi{Q}(y|i)} \Big) \nonumber \\
&\overset{(a)}{\le}  \sum_{j=1}^{k/2} 
\max_{\nu:\nu_j=-1} \sum_{y\in\cY} \Big( \Big(\sum_{i=1}^k \mathbi{p}_{\nu}(i)\mathbi{Q}(y|i) \Big)
\frac{\sum_{i=1}^k \mathbi{p}_{\nu}(i)\mathbi{Q}(y|i)- \sum_{i=1}^k \mathbi{p}_{\nu+2e_j}(i)\mathbi{Q}(y|i)}
{\sum_{i=1}^k \mathbi{p}_{\nu+2e_j}(i)\mathbi{Q}(y|i)} \Big) \nonumber \\
&\overset{(b)}{=}  \sum_{j=1}^{k/2} 
\max_{\nu:\nu_j=-1} \sum_{y\in\cY} 
\frac{\left(  \sum_{i=1}^k \mathbi{p}_{\nu}(i)\mathbi{Q}(y|i)- \sum_{i=1}^k \mathbi{p}_{\nu+2e_j}(i)\mathbi{Q}(y|i)  \right)^2}
{\sum_{i=1}^k \mathbi{p}_{\nu+2e_j}(i)\mathbi{Q}(y|i)} \nonumber \\
&= \sum_{j=1}^{k/2} 
\max_{\nu:\nu_j=-1} \sum_{y\in\cY} 
\frac{\Big(  \frac{2\delta}{k} \mathbi{Q}(y|j+k/2) -  \frac{2\delta}{k} \mathbi{Q}(y|j)   \Big)^2}
{\sum_{i=1}^k \mathbi{p}_{\nu+2e_j}(i)\mathbi{Q}(y|i)} \nonumber \\
&\overset{(c)}{\le}  \sum_{j=1}^{k/2}  \sum_{y\in\cY} 
\frac{\Big(  \frac{2\delta}{k} \mathbi{Q}(y|j+k/2) -  \frac{2\delta}{k} \mathbi{Q}(y|j)   \Big)^2}
{\frac{1}{k}(1-\delta)\sum_{i=1}^k \mathbi{Q}(y|i)} \nonumber \\
&= \frac{4\delta^2}{k(1-\delta)} \sum_{j=1}^{k/2} \sum_{y\in\cY}
\Big( (\sum_{i=1}^k \mathbi{Q}(y|i)) \Big(
\frac{  \mathbi{Q}(y|j+k/2) -   \mathbi{Q}(y|j)  }
{\sum_{i=1}^k \mathbi{Q}(y|i)}   \Big)^2 \Big) \nonumber \\
&= \frac{4\delta^2}{k(1-\delta)}  \sum_{y\in\cY}
\Big( (\sum_{i=1}^k \mathbi{Q}(y|i)) 
\sum_{j=1}^{k/2} \Big(
\frac{  \mathbi{Q}(y|j+k/2) -   \mathbi{Q}(y|j)  }
{\sum_{i=1}^k \mathbi{Q}(y|i)}   \Big)^2 \Big) \nonumber \\
&\le  \frac{4\delta^2}{k(1-\delta)} \Big( \sum_{y\in\cY}
 \sum_{i=1}^k \mathbi{Q}(y|i)
 \Big) \max_{y\in\cY} \sum_{j=1}^{k/2} \Big(
\frac{  \mathbi{Q}(y|j+k/2) -   \mathbi{Q}(y|j)  }
{\sum_{i=1}^k \mathbi{Q}(y|i)}   \Big)^2 \nonumber \\
&= \frac{4\delta^2}{1-\delta} \max_{y\in\cY} \sum_{j=1}^{k/2} \Big(
\frac{  \mathbi{Q}(y|j+k/2) -   \mathbi{Q}(y|j)  }
{\sum_{i=1}^k \mathbi{Q}(y|i)}   \Big)^2 \nonumber \\
&\overset{(d)}{\le} 
 {\begin{cases}
    \displaystyle\frac{4\delta^2}{1-\delta} \frac{2(e^{\epsilon}-1)^2}{k(e^{\epsilon}+1)^2} & \text{if }  e^{\epsilon}<3 \\
\displaystyle\frac{4\delta^2}{1-\delta}
 \frac{e^{\epsilon}-1}{4k} & \text{if }  e^{\epsilon}\ge 3
   \end{cases}} ,\label{eq:lengthy}
     \end{align}
 where $(a)$ follows from the fact that $\log(x)\le x-1$ for all $x>0;$ $(b)$ follows from the normalization
   $$
   \sum_{y\in\cY} \sum_{i=1}^k \mathbi{p}_{\nu}(i)\mathbi{Q}(y|i)
=\sum_{y\in\cY} \sum_{i=1}^k \mathbi{p}_{\nu+2e_j}(i)\mathbi{Q}(y|i) = 1;
$$
$(c)$ follows from the fact that $\mathbi{p}_{\nu}(i) \ge (1-\delta)/k$ for all $\nu\in\cV$ and all $i=1,2,\dots,k,$ and finally, $(d)$ follows from Lemma~\ref{lemimp} below.

Substituting \eqref{eq:lengthy} into \eqref{eq:bdrq}, we obtain that for every 
$\mathbi{Q}\in \cD_{\epsilon,E},$
    \begin{equation}\label{eq:fin}
     \begin{aligned}
r_{k,n}^{\ell_2^2} (\mathbi{Q}) & \ge
\frac{\delta^2}{4k} \Big(1- \sqrt{ \frac{\delta^2}{1-\delta} 
\frac{8n(e^{\epsilon}-1)^2}{k^2(e^{\epsilon}+1)^2} } \Big)
\text{~for~} e^{\epsilon}<3, \\
r_{k,n}^{\ell_2^2} (\mathbi{Q}) & \ge
\frac{\delta^2}{4k} \Big(1- \sqrt{ \frac{\delta^2}{1-\delta} 
 \frac{n(e^{\epsilon}-1)}{k^2} } \Big)
\text{~for~} e^{\epsilon}\ge 3 
   \end{aligned}
   \end{equation}
For $e^{\epsilon}<3,$ let $\delta^2= \frac{k^2(e^{\epsilon}+1)^2}{64n(e^{\epsilon}-1)^2},$
we have 
$$
r_{k,n}^{\ell_2^2} (\mathbi{Q}) \overset{(a)}{\ge} \frac{k(e^{\epsilon}+1)^2}{512n(e^{\epsilon}-1)^2}
\overset{(c)}{\ge}  \frac{(k-1)(e^{\epsilon}+1)^2}{512n(e^{\epsilon}-1)^2}.
$$
For $e^{\epsilon}\ge3,$ let $\delta^2=\frac{k^2}{8n(e^{\epsilon}-1)},$ we have
$$
r_{k,n}^{\ell_2^2} (\mathbi{Q}) \overset{(b)}{\ge} \frac{k}{64n(e^{\epsilon}-1)}
\overset{(d)}{\ge}  \frac{k-1}{64n(e^{\epsilon}-1)},
$$
where $(a)$ and $(b)$ follows from the condition $n>\max(\frac{k^2(e^{\epsilon}+1)^2}{16(e^{\epsilon}-1)^2}, \frac{k^2}{2(e^{\epsilon}-1)}).$ This condition guarantees that
$1-\sqrt{\frac{1}{8(1-\delta)}}\ge \frac{1}{2}.$ 
The inequalities $(c)$ and $(d)$ are for the purpose of giving unified lower bounds for both even and odd $k.$
This completes the proof for $\ell=\ell_2^2.$

\subsubsection{Loss function $\ell_1$}
The proof for $\ell=\ell_1$ is very similar to the proof above. The only difference is that in equation \eqref{eq:initmul} we have $\delta/k$ instead of $\delta^2/k^2$ as the constant on the right-hand side:
$$
\ell_1(\hat{\mathbi{p}}(y^n), \mathbi{p}_{\nu})
\ge \sum_{j=1}^{k/2} \left|\hat{p}_j(y^n)-(\frac{1}{k} + \frac{\delta \nu_j}{k}) \right|
\ge \frac{\delta}{k} \sum_{j=1}^{k/2} \mathbbm{1}\{\sign(\hat{p}_j(y^n)-\frac{1}{k}) \neq \nu_j\}.
$$
Parallelling \eqref{eq:fin}, we can show that for every 
$\mathbi{Q}\in \cD_{\epsilon,E},$
\begin{align*}
r_{k,n}^{\ell_1} (\mathbi{Q}) & \ge
\frac{\delta}{4} \Big(1- \sqrt{ \frac{\delta^2}{1-\delta} 
\frac{8n(e^{\epsilon}-1)^2}{k^2(e^{\epsilon}+1)^2} } \Big)
\text{~for~} e^{\epsilon}<3, \\
r_{k,n}^{\ell_1} (\mathbi{Q}) & \ge
\frac{\delta}{4} \Big(1- \sqrt{ \frac{\delta^2}{1-\delta} 
 \frac{n(e^{\epsilon}-1)}{k^2} } \Big)
\text{~for~} e^{\epsilon}\ge 3 
\end{align*}
For $e^{\epsilon}<3,$ taking $\delta^2= \frac{k^2(e^{\epsilon}+1)^2}{64n(e^{\epsilon}-1)^2},$
we have 
$$
r_{k,n}^{\ell_1} (\mathbi{Q}) \ge \frac{k(e^{\epsilon}+1)}{64\sqrt{n}(e^{\epsilon}-1)}
\overset{(a)}{\ge}  \frac{(k-1)(e^{\epsilon}+1)}{64\sqrt{n}(e^{\epsilon}-1)}.
$$
For $e^{\epsilon}\ge3,$ taking $\delta^2=\frac{k^2}{8n(e^{\epsilon}-1)},$ we have
$$
r_{k,n}^{\ell_1} (\mathbi{Q}) \ge \frac{k}{16\sqrt{2n(e^{\epsilon}-1)}}
\overset{(b)}{\ge}  \frac{k-1}{16\sqrt{2n(e^{\epsilon}-1)}}.
$$
Similarly, the inequalities $(a)$ and $(b)$ above are for the purpose of giving unified lower bounds for both even and odd $k.$
This completes the proof for $\ell=\ell_1.$

For odd $k,$ the only change we need to make in this proof is to set $\cV=\{-1,1\}^{(k-1)/2},$ and for $\nu\in\cV$ let $\mathbi{p}_{\nu}$ be the distribution
$$
\mathbi{p}_{\nu} := \frac{1}{k-1} \left[\begin{array}{@{}c@{}}\mathbf{1}_{k-1} \\ 0 \end{array}\right] + \frac{\delta}{k-1}
\left[\begin{array}{@{}c@{}}\nu \\ -\nu \\ 0 \end{array}\right]
\in\Delta_k,
$$
where $\mathbf{1}_{k-1}$ is the all $1$ vector with length $k-1.$
The rest of the proof is exactly the same as the proof for even $k.$
%\end{IEEEproof}

\begin{lemma}\label{lemimp}
If $k$ is even, and $\mathbi{Q}\in\cD_{\epsilon,E},$ then for all $y\in\cY$
$$
\sum_{j=1}^{k/2} \Big(
\frac{  \mathbi{Q}(y|j+k/2) -   \mathbi{Q}(y|j)  }
{\sum_{i=1}^k \mathbi{Q}(y|i)}   \Big)^2 
\le \begin{cases}
\displaystyle\frac{2(e^{\epsilon}-1)^2}{k(e^{\epsilon}+1)^2} & \mbox{if }  e^{\epsilon}<3 \\
 \displaystyle\frac{e^{\epsilon}-1}{4k} & \mbox{if }  e^{\epsilon}\ge 3.
\end{cases}
 $$
\end{lemma}
\begin{IEEEproof}
Let $\tilde{\mathbi{Q}}(y|i)=\mathbi{Q}(y|i)/(\min_{x\in\cX}\mathbi{Q}(y|x)).$ Since $\mathbi{Q}\in\cD_{\epsilon,E},$ we have $\tilde{\mathbi{Q}}(y|i)=1$ or $e^{\epsilon}$ for all $y\in\cY$ and $i\in\cX.$ It is also clear that 
$$
\sum_{j=1}^{k/2} \Big(
\frac{  \mathbi{Q}(y|j+k/2) -   \mathbi{Q}(y|j)  }
{\sum_{i=1}^k \mathbi{Q}(y|i)}   \Big)^2 
=
\sum_{j=1}^{k/2} \Big(
\frac{  \tilde{\mathbi{Q}}(y|j+k/2) -  \tilde{ \mathbi{Q}}(y|j)  }
{\sum_{i=1}^k \tilde{\mathbi{Q}} (y|i)}   \Big)^2.
$$
We would like to find a vector $(\tilde{\mathbi{Q}} (y|1), \tilde{\mathbi{Q}} (y|2), \dots, \tilde{\mathbi{Q}} (y|k)) \in \{1,e^{\epsilon}\}^k$ that maximizes the right-hand side of the last equation.
First observe that if $\tilde{\mathbi{Q}}(y|j+k/2) =  \tilde{ \mathbi{Q}}(y|j)=e^{\epsilon}$ for some $j\in\{1,2,\dots,k/2\},$ then resetting $\tilde{ \mathbi{Q}}(y|j)=1$ increases the numerator and decreases the denominator, and thus increases the value of the expression above. As a result, in order to maximize the expression above, at least one of the two numbers $\tilde{\mathbi{Q}}(y|j+k/2)$ and  $\tilde{ \mathbi{Q}}(y|j)$ must be $1$ for all $j=1,2,\dots,k/2.$ Under this condition, we have 
$t :=|\{i\in\{1,2,\dots,k\}: \tilde{\mathbi{Q}} (y|i)=e^{\epsilon}\}| \le k/2.$ Moreover,
$$
\sum_{j=1}^{k/2} \Big(
\frac{  \tilde{\mathbi{Q}}(y|j+k/2) -  \tilde{ \mathbi{Q}}(y|j)  }
{\sum_{i=1}^k \tilde{\mathbi{Q}} (y|i)}   \Big)^2
=\frac{t(e^{\epsilon}-1)^2}{(t(e^{\epsilon}-1)+k)^2}.
$$
We want to choose $t\in\{0,1,\dots,k/2\}$ to maximize the expression above. It is clear that $t=0$ does not maximize this expression, thus we can restrict ourselves to $t\in\{1,2,\dots,k/2\}.$ We have
\begin{equation} \label{eq:justify}
\frac{t(e^{\epsilon}-1)^2}{(t(e^{\epsilon}-1)+k)^2}
=\frac{1}{(\sqrt{t}+\frac{k}{\sqrt{t}(e^{\epsilon}-1)} )^2}.
%\le \left\{ \begin{array}{ccc}
%\frac{2(e^{\epsilon}-1)^2}{k(e^{\epsilon}+1)^2} & \mbox{if} & e^{\epsilon}<3, \\
% \frac{e^{\epsilon}-1}{4k} & \mbox{if} &  e^{\epsilon}\ge 3.
%\end{array} \right.
\end{equation}
The right-hand side of \eqref{eq:justify} can be easily seen to satisfy the inequalities in the statement of the lemma.
\end{IEEEproof}

\subsection{Asymptotic behavior of the $\ell_2^2$ and $\ell_1$ risk}
In this part we derive the asymptotic behavior of the $\ell_2^2$ and $\ell_1$ risk.

\begin{theorem}\label{Thm:last} Let $e^{\epsilon} \ll k,$ then for $n$ large enough,
$$
r_{\epsilon, k,n}^{\ell_2^2} =  \Theta \Big( \frac{k e^{\epsilon}}{n (e^{\epsilon}-1)^2} \Big)   , \quad
r_{\epsilon, k,n}^{\ell_1}  =     \Theta \Big(\frac{k \sqrt{ e^{\epsilon}}}{(e^{\epsilon}-1) \sqrt{n}} \Big) .
$$
\end{theorem}
\begin{IEEEproof}
According to Theorem~\ref{Thm:lb}, for $e^{\epsilon}<3,$
\begin{align*}
r_{\epsilon, k,n}^{\ell_2^2}  \ge  \frac{(k-1)(e^{\epsilon}+1)^2}{512n(e^{\epsilon}-1)^2}
\ge  \frac{k e^{\epsilon}}{512n(e^{\epsilon}-1)^2 } \Big(1-\frac{1}{k} \Big)
=  \Theta \Big(\frac{k e^{\epsilon}}{n (e^{\epsilon}-1)^2} \Big), \\
r_{\epsilon, k,n}^{\ell_1}  \ge \frac{(k-1)(e^{\epsilon}+1)}{64\sqrt{n}(e^{\epsilon}-1)}
\ge  \frac{k \sqrt{e^{\epsilon}} }{64\sqrt{n}(e^{\epsilon}-1)} \Big(1-\frac{1}{k} \Big)
= \Theta \Big(\frac{k \sqrt{ e^{\epsilon}}}{(e^{\epsilon}-1) \sqrt{n}} \Big).
\end{align*}
For $e^{\epsilon}\ge 3,$
\begin{align*}
r_{\epsilon, k,n}^{\ell_2^2}  \ge \frac{k-1}{64n(e^{\epsilon}-1)}
& \ge  \frac{ k e^{\epsilon}}{96 n(e^{\epsilon}-1)^2 } \Big(1-\frac{1}{k} \Big)
= \Theta \Big(\frac{k e^{\epsilon}}{n (e^{\epsilon}-1)^2} \Big) , \\
r_{\epsilon, k,n}^{\ell_1}  \ge \frac{k-1}{16\sqrt{2n(e^{\epsilon}-1)}}
& \ge \frac{k}{16}\sqrt{\frac{1}{2n(e^{\epsilon}-1)} } \sqrt{\frac{2 e^{\epsilon}}{3(e^{\epsilon}-1)} }
\Big(1-\frac{1}{k} \Big)
=     \Theta\Big( \frac{k \sqrt{ e^{\epsilon}}}{(e^{\epsilon}-1) \sqrt{n}} \Big) .
\end{align*}
Combined with \eqref{eq:orderub}, this completes the proof.
\end{IEEEproof}

\begin{remark}
When $\epsilon$ is close to $0,$ Theorem~\ref{Thm:last} implies that 
$
r_{\epsilon, k,n}^{\ell_2^2} =  \Theta(\frac{k }{n \epsilon^2})   ,
r_{\epsilon, k,n}^{\ell_1}  =     \Theta(\frac{k }{\epsilon \sqrt{n}} ) ,
$
which coincides with the bounds given in \cite{Duchi16} in this regime, as expected.
\end{remark}

\begin{remark}
In \eqref{eq:justify}, the left-hand side takes the maximum value for $t=k/(e^{\epsilon}-1).$ Note that the parameter 
$t$ here plays the same role as the parameter $d$ in Section~\ref{Sect:new}. This gives some intuition why $d\approx k/e^{\epsilon}$ is optimal.
\end{remark}

\begin{remark}
The main technical improvement over \cite{Duchi16} in the proof of the lower bound in this section is the bound in Lemma~\ref{lemimp}. 
In \cite{Duchi16}, the authors bound the numerator and denominator separately. 
They bound the denominator in the following straightforward way:
$\sum_{i=1}^k \mathbi{Q}(y|i) \ge k (\min_{x\in\cX} \mathbi{Q}(y|x) ).$ Their method leads to a tight bound only when $e^{\epsilon}$ is very close to $1,$ because only in this case their bound on the denominator is tight. Our method in Lemma~\ref{lemimp}, on the other hand, treats the numerator and denominator as a whole, and leads to a tight lower bound for the much larger region $e^{\epsilon} \ll k.$
\end{remark}

\appendices
\section{Proof of Proposition \ref{prop:risks}}\label{ap:risk}
%\begin{IEEEproof}
First let us check that the estimator $\hat{\mathbi{p}}$ in \eqref{eq:emp} is unbiased.
We have $\underset{Y^n\sim \mathbi{m}^n}{\mathbb{E}}\frac{T_i}{n}= \mathbi{m}(Y_i = 1)=q_i$ (see \eqref{eq:Py}), and so
$$
\underset{Y^n\sim \mathbi{m}^n}{\mathbb{E}}  \hat{p_i}=
 \Big(\frac{(k-1)e^{\epsilon}+\frac{(k-1)(k-d)}{d}}{(k-d)(e^{\epsilon}-1)}\Big)
\underset{Y^n\sim \mathbi{m}^n}{\mathbb{E}} \frac{T_i}{n}
-\frac{(d-1)e^{\epsilon}+k-d}{(k-d)(e^{\epsilon}-1)}=p_i.
$$
To shorten the formulas, let $K:=\frac{(k-1)e^{\epsilon}+{(k-1)(k-d)}/{d}}{(k-d)(e^{\epsilon}-1)}.$ We have
\begin{align*}
\underset{Y^n\sim \mathbi{m}^n}{\mathbb{E}} 
\ell_2^2(\hat{\mathbi{p}}(Y^n),\mathbi{p})&=
\sum_{i=1}^k \underset{Y^n\sim \mathbi{m}^n}{\mathbb{E}} (\hat{p_i} - p_i)^2 
= \sum_{i=1}^k \underset{Y^n\sim \mathbi{m}^n}{\mathbb{E}} \left( K\frac{T_i}{n}
-\frac{(d-1)e^{\epsilon}+k-d}{(k-d)(e^{\epsilon}-1)} - p_i \right)^2 \\
&= \sum_{i=1}^k \underset{Y^n\sim \mathbi{m}^n}{\mathbb{E}} \left( K\frac{T_i}{n}
- K
\underset{Y^n\sim \mathbi{m}^n}{\mathbb{E}} \left(\frac{T_i}{n}\right) \right)^2 \\
&=  K^2
 \sum_{i=1}^k \underset{Y^n\sim \mathbi{m}^n}{\mathbb{E}} \left( \frac{T_i}{n}
- \underset{Y^n\sim \mathbi{m}^n}{\mathbb{E}} \left(\frac{T_i}{n}\right) \right)^2 \\
&=  \frac{K^2}{n} 
 \sum_{i=1}^k {\Var} (Y_i)
{=} \frac{K^2}{n} 
 \sum_{i=1}^k q_i(1-q_i).
\end{align*}
Now substitute $q_i$ from \eqref{eq:Py}:
\begin{align*} 
\underset{Y^n\sim \mathbi{m}^n}{\mathbb{E}} 
\ell_2^2( & \hat{\mathbi{p}}(Y^n), \mathbi{p})= \frac{K^2}{n}
\biggl(\sum_{i=1}^k \left(\frac{(k-d)(e^{\epsilon}-1)p_i+(d-1)e^{\epsilon}+k-d}
{(k-1)e^{\epsilon}+\frac{(k-1)(k-d)}{d}} \right) \\
   &\hspace*{2in}- \sum_{i=1}^k \left(\frac{(k-d)(e^{\epsilon}-1)p_i+(d-1)e^{\epsilon}+k-d}
{(k-1)e^{\epsilon}+\frac{(k-1)(k-d)}{d}} \right)^2 \biggr) \\
\overset{(a)}{=} & \frac{K^2}{n}
\left( \frac{(k-d)(e^{\epsilon}-1)+k \big( (d-1)e^{\epsilon}+k-d \big)}
{(k-1)e^{\epsilon}+\frac{(k-1)(k-d)}{d}} \right. \\
& \left. -  \frac{(k-d)^2(e^{\epsilon}-1)^2 \sum_{i=1}^k p_i^2+k ((d-1)e^{\epsilon}+k-d)^2
+2(k-d)(e^{\epsilon}-1) \big( (d-1)e^{\epsilon}+k-d \big) }
{\big( (k-1)e^{\epsilon}+\frac{(k-1)(k-d)}{d} \big)^2 }  \right) \\
= & \frac{1}{n}
\left( \frac{\Big( (k-d)(e^{\epsilon}-1)+k \big( (d-1)e^{\epsilon}+k-d \big) \Big) 
 \Big( (k-1)e^{\epsilon}+\frac{(k-1)(k-d)}{d} \Big) }
{(k-d)^2(e^{\epsilon}-1)^2} \right. \\
& \left. -  \frac{ k ((d-1)e^{\epsilon}+k-d)^2
+2(k-d)(e^{\epsilon}-1) \big( (d-1)e^{\epsilon}+k-d \big) }
{(k-d)^2(e^{\epsilon}-1)^2 } -\sum_{i=1}^k p_i^2 \right) \\
= & \frac{1}{n}
\left( \frac{d(k-1)^2 e^{2\epsilon} +2(k-d)(k-1)^2 e^{\epsilon} +\frac{(k-1)^2(k-d)^2}{d}}
{(k-d)^2(e^{\epsilon}-1)^2} \right. \\
& \left. -  \frac{ (d-1) (kd+k-2d) e^{2\epsilon} + 2(k-d)(kd-2d+1) e^{\epsilon} +(k-2)(k-d)^2 }
{(k-d)^2(e^{\epsilon}-1)^2 } -\sum_{i=1}^k p_i^2 \right) \\
= & \frac{1}{n}
\left(\frac{(d(k-2)+1)e^{2\epsilon}}{(k-d)(e^{\epsilon}-1)^2} +\frac{2(k-2)e^{\epsilon}}{(e^{\epsilon}-1)^2}
+\frac{(k-2)(k-d)+1}{d(e^{\epsilon}-1)^2} -\sum_{i=1}^k p_i^2 \right),
\end{align*}
where for $(a)$ we use $\sum_{i=1}^k p_i=1.$
This proves \eqref{eq:L2}. 

Similarly, for $\ell_1$ risk,
  \begin{align*}
\underset{Y^n\sim \mathbi{m}^n}{\mathbb{E}} \ell_1(\hat{\mathbi{p}}(Y^n),\mathbi{p}) &=
\sum_{i=1}^k \underset{Y^n\sim \mathbi{m}^n}{\mathbb{E}} |\hat{p_i} - p_i| \\
&= \sum_{i=1}^k \underset{Y^n\sim \mathbi{m}^n}{\mathbb{E}} \left| K\frac{T_i}{n}
-\frac{(d-1)e^{\epsilon}+k-d}{(k-d)(e^{\epsilon}-1)} - p_i \right| \\
&= \sum_{i=1}^k \underset{Y^n\sim \mathbi{m}^n}{\mathbb{E}} \left| K \frac{T_i}{n}
- K
\underset{Y^n\sim \mathbi{m}^n}{\mathbb{E}} \left(\frac{T_i}{n}\right) \right| \\
&=  K
 \sum_{i=1}^k \underset{Y^n\sim \mathbi{m}^n}{\mathbb{E}} \left| \frac{T_i}{n}
- \underset{Y^n\sim \mathbi{m}^n}{\mathbb{E}} \left(\frac{T_i}{n}\right) \right| \\
&=  \frac{1}{\sqrt{n}} K
 \sum_{i=1}^k \underset{Y^n\sim \mathbi{m}^n}{\mathbb{E}} \left| 
\frac{T_i - \underset{Y^n\sim \mathbi{m}^n}{\mathbb{E}} T_i}{\sqrt{n}} \right|.  
  \end{align*}
By the central limit theorem we now claim that for $n\to\infty$ the RV $\frac{T_i - {\mathbb{E}} T_i}{\sqrt{n}}$ converges in distribution to a Gaussian RV $Z\sim \cN(0,q_i(1-q_i))$
As a result,
$$
\lim_{n\to\infty} \underset{Y^n\sim \mathbi{m}^n}{\mathbb{E}} \left| 
\frac{T_i - \underset{Y^n\sim \mathbi{m}^n}{\mathbb{E}} T_i}{\sqrt{n}} \right|
= {\mathbb{E}} |Z| = \sqrt{\frac{2}{\pi}q_i(1-q_i)}.
$$
Therefore, for large $n$ we continue as follows
  \begin{align*}
 &\underset{Y^n\sim \mathbi{m}^n}{\mathbb{E}} \ell_1(\hat{\mathbi{p}}(Y^n),\mathbi{p}) =
 \frac{K}{\sqrt{n}}  
\sum_{i=1}^k \sqrt{\frac{2}{\pi}q_i(1-q_i)} + o\Big(\frac 1{\sqrt n}\Big) \\
= & K \sum_{i=1}^k \sqrt{\frac{2}{\pi n} \left(\frac{(k-d)(e^{\epsilon}-1)p_i+(d-1)e^{\epsilon}+k-d}
{(k-1)e^{\epsilon}+\frac{(k-1)(k-d)}{d}}- \left( \frac{(k-d)(e^{\epsilon}-1)p_i+(d-1)e^{\epsilon}+k-d}
{(k-1)e^{\epsilon}+\frac{(k-1)(k-d)}{d}} \right)^2 \right)} + o\Big(\frac 1{\sqrt n}\Big)\\
= & \frac{1}{e^{\epsilon}-1} \sum_{i=1}^k \sqrt{\frac{2}{\pi n} \Big( (e^{\epsilon}-1)p_i+ \frac{(d-1)e^{\epsilon}}{k-d}+1 \Big)
\Big((e^{\epsilon}-1)(1-p_i)+\frac{k-1}{d} \Big)} + o\Big(\frac 1{\sqrt n}\Big).
\end{align*}

\section{Proof of Proposition \ref{prop:05}}\label{ap:05}
%\begin{IEEEproof}
From \eqref{eq:longub} we have
$$
r_{\OPT}^{\ell_2^2}
< \frac{4k e^{\epsilon}}{n (e^{\epsilon}-1)^2}
\Big(1 + \frac{2e^{\epsilon}+3}{4k}\Big).
$$
According to \eqref{eq:RAPrisk},
$$
r_{\RAP}^{\ell_2^2} \ge  r_{\RAP}^{\ell_2^2}(\mathbi{p}_U)
> \frac{k e^{\epsilon /2}}{n (e^{\epsilon/2}-1)^2 }.
$$
$r_{\RR}^{\ell_2^2}(\mathbi{p})$ is given in \eqref{eq:L2} upon plugging in $d=1.$ So we have
$$
r_{\RR}^{\ell_2^2} \ge r_{\RR}^{\ell_2^2}(\mathbi{p}_U)
> \frac{k^2}{n (e^{\epsilon}-1)^2} \Big(1-\frac{1}{k}\Big).
$$
Using the conditions on $\epsilon$ and $k$ in the statement we can easily calculate that
\begin{align*}
r_{\OPT}^{\ell_2^2} / r_{\RAP}^{\ell_2^2} < \frac{4 e^{\epsilon/2} }{(e^{\epsilon/2}+1)^2 }
\Big(1 + \frac{2e^{\epsilon}+3}{4k}\Big) <1/2 , \\
r_{\OPT}^{\ell_2^2} / r_{\RR}^{\ell_2^2} < \frac{4e^{\epsilon}}{k} \Big(1 + \frac{2e^{\epsilon}+3}{4k}\Big)
\Big(1-\frac{1}{k}\Big)^{-1} <1/2.
\end{align*}
For large $n,$ according to \eqref{eq:l1ub},
$$
r_{\OPT}^{\ell_1} <
\sqrt{\frac{8}{\pi }}  \frac{k \sqrt{ e^{\epsilon}}}{(e^{\epsilon}-1) \sqrt{n}} 
 \Big(1+\frac{e^{\epsilon}+1}{4k}\Big) .
$$
According to \eqref{eq:RAPrisk},
$$
r_{\RAP}^{\ell_1} \ge \max_{\mathbi{p}\in\Delta_k}  r_{\RAP}^{\ell_1}(\mathbi{p}_U)
>  \sqrt{ \frac{2}{\pi } } \frac{k \sqrt{ e^{\epsilon/2}}}{(e^{\epsilon/2}-1)\sqrt{n}}  \Big(1-\frac{1}{k}\Big)^{1/2}.
$$
The quantity $r_{\RR}^{\ell_1}(\mathbi{p})$ is given by \eqref{eq:L1} once we take $d=1$ in it. We obtain
$$
r_{\RR}^{\ell_1} \ge r_{\RR}^{\ell_1}(\mathbi{p}_U)
> \sqrt{\frac{2}{\pi } } \frac{k \sqrt{k}}{(e^{\epsilon}-1) \sqrt{n}}  \Big(1-\frac{1}{k}\Big)^{1/2}.
$$
Therefore,
\begin{align*}
r_{\OPT}^{\ell_1} / r_{\RAP}^{\ell_1} < \frac{2 e^{\epsilon/4}}{e^{\epsilon/2}+1}
\Big(1+\frac{e^{\epsilon}+1}{4k}\Big)  \Big(1-\frac{1}{k}\Big)^{-1/2}  < 0.7, \\
r_{\OPT}^{\ell_1} / r_{\RR}^{\ell_1} < 2\sqrt{\frac{e^{\epsilon} }{k} }  \Big(1+\frac{e^{\epsilon}+1}{4k}\Big) 
 \Big(1-\frac{1}{k}\Big)^{-1/2}  < 0.7.
\end{align*}
%\end{IEEEproof}

\section{Proof of Lemma \ref{lem:tech}}\label{ap:approx}
Consider the set of {\em measurable rectangles} 
$$
\cR=\{C_1\times C_2\times \dots \times C_n: C_i\in \sigma(\cY), i=1,2,\dots,n\}.
$$
The $n$-fold product $\sigma$-algebra $\sigma(\cY)^{\times n}$ is the algebra generated by the set $\cR$ of measurable rectangles. With a mild abuse of notation 
we will write $\sigma(\cR)$ instead of $\sigma(\cY)^{\times n}.$ 
The product measure is the Carath\'eodory extension of the premeasure on $\cR$ \cite[Ch.~20]{Royden10}. 
More specifically, for any multi-index $\underline{j}=(j_1,j_2,\dots,j_n) \in \{1,2,\dots,k\}^n,$ the premeasure $\lambda_{\underline{j}}: \cR \to [0,1]$ is defined by
$$
\lambda_{\underline{j}}(C_1\times C_2\times \dots \times C_n)
=\prod_{i=1}^n P_{j_i}(C_i) \text{~for~} C_1\times C_2\times \dots \times C_n \in \cR,
$$
and $P_{\underline{j}}$ is the extension of $\lambda_{\underline{j}}: \cR \to [0,1]$ on the $\sigma$-algebra $\sigma(\cR).$

(a) {\em Definition of the sets $B_i'$}.
Since $B_i \in \sigma(\cR),i=1,2,\dots,N$ and since the product measure is $\sigma$-finite, we can use the Carath\'eodory-Hahn theorem
to claim that for every $i$ and every $\underline{j} \in [k]^n,$
there exists a countable collection  of sets $\{C_j^{i;\underline{j}}\}_{j=1}^{\infty}\in\cR$ such that \cite[p.~353]{Royden10}
\begin{equation}\label{eq:Bi}
B_i \subseteq \bigcup_{j=1}^{\infty} C_j^{i;\underline{j}} \text{~and~}
\sum_{j=1}^{\infty} P_{\underline{j}}(C_j^{i;\underline{j}})
< P_{\underline{j}} (B_i) + \frac{\alpha}{N^2}.
\end{equation}
Define 
  \begin{equation}\label{eq:Cij}
B_i''=\bigcap_{\underline{j}\in[k]^n} \bigcup_{j=1}^{\infty} C_j^{i;\underline{j}} .
  \end{equation}
By \eqref{eq:Bi}, 
we have 
\begin{equation}\label{eq:dbprime}
B_i\subseteq B_i'', \text{~and~}
 P_{\underline{j}}(B_i'')
< P_{\underline{j}} (B_i) + \frac{\alpha}{N^2} \text{ for all } \underline{j}\in[k]^n.
\end{equation}
Therefore,  for all $\underline{j}\in[k]^n,$
\begin{equation}\label{eq:setdiff}
 P_{\underline{j}}(B_i''\setminus B_i)
<  \frac{\alpha}{N^2}.
\end{equation}
Next we would like to write the sets $B_i''$ as countable unions of sets. To this end, we interchange the union and intersection
in \eqref{eq:Cij} and deduce that $B_i''$ is the union of the following countable collection of sets
$$
\cC_i=\Big\{\bigcap_{\underline{j}\in [n]^k} C_{N_{\underline{j}}}^{i;\underline{j}}:
N_{\underline{j}} \in \mathbb{N} \text{~for all~} \underline{j}\in[k]^n \Big\}.
$$
%Here the natural number $N_{\underline{j}}$ corresponds 
Since finite intersections of measurable rectangles are still measurable rectangles, $\cC_i\subseteq \cR.$
We re-label all the sets in $\cC_i$ as 
$\cC_i=\{C^{i,j}\}_{j=1}^{\infty},$ where $C^{i,j}\in\cR$ for all $j\ge 1.$ Thus,
$$
B_i''=\bigcup_{C\in\cC_i}C=\bigcup_{j=1}^{\infty}C^{i,j}.
$$
By continuity of measure, for every $\underline{j}\in[k]^n$ there exists a sufficiently large $N_{i;\underline{j}}$ such that 
$$
P_{\underline{j}} \biggl(\bigcup_{j=1}^{N_{i;\underline{j}}}C^{i,j}\biggr) >
P_{\underline{j}} (B_i'') - \frac{\alpha}{N^2}.
$$
Let $N_i=\max_{\underline{j}\in[k]^n}N_{i;\underline{j}}.$ 
Then for all $\underline{j}\in[k]^n$
\begin{equation}\label{eq:Ni}
P_{\underline{j}} \biggl(\bigcup_{j=1}^{N_i}C^{i,j}\biggl) >  
P_{\underline{j}} (B_i'') - \frac{\alpha}{N^2}.
\end{equation}
Now let us define the sets $B_i'$ whose existence is claimed in the statement of the lemma:
\begin{equation}\label{eq:defBi}
B_i'=
\begin{cases}\bigcup_{j=1}^{N_1}C^{1,j} &\text{if } i=1,\\[.05in]
\Big(\bigcup_{j=1}^{N_i}C^{i,j}\Big)\setminus \Big(\bigcup_{u=1}^{i-1} B_u'\Big) &\text{if~} i=2,3,\dots, N-1,\\[.05in]
\cY^n \setminus \Big(\bigcup_{i=1}^{N-1} B_i'\Big) &\text{if }i=N.
\end{cases}
\end{equation}
By definition these sets form a partition of $\cY^n.$

\vspace*{.05in}
(b) {\em Proof that the sets $B_i'$ satisfy \eqref{eq:alpha}.}
%Now we prove that condition (3) in the lemma is satisfied by this choice of $\{B_i'\}_{i=1}^N$. 
Let $i=1.$ Since $B_1'\subseteq B_1''$ and $B_1\subseteq B_1'',$ we can use \eqref{eq:dbprime} and \eqref{eq:Ni} to claim
that for all $\underline{j}\in[k]^n$
$$
| P_{\underline{j}}(B_1')
- P_{\underline{j}} (B_1) | < \frac{\alpha}{N^2}.
$$
For $i=2,3,\dots,N-1,$ since $B_i'\subseteq B_i'',$ for all $\underline j\in[k]^n$ we obtain from \eqref{eq:dbprime},
\begin{equation}\label{eq:oneside}
 P_{\underline{j}}(B_i') 
- P_{\underline{j}} (B_i) < \frac{\alpha}{N^2}
\le \frac{\alpha}{N}.
\end{equation}
By definition,
\begin{align*}
B_i\setminus B_i' &= B_i \cap \Big( \big(\bigcup_{j=1}^{N_i}C^{i,j}\big)\cap
\big( \bigcup_{l=1}^{i-1}B_l' \big)^c  \Big)^c\\
&= B_i \cap \Big( \big(\bigcup_{j=1}^{N_i}C^{i,j}\big)^c \cup
\big( \bigcup_{l=1}^{i-1}B_l' \big)  \Big)\\
&= \Big( B_i \cap  \big(\bigcup_{j=1}^{N_i}C^{i,j}\big)^c \Big) \cup
\Big( B_i \cap \big( \bigcup_{l=1}^{i-1}B_l' \big) \Big)\\
&= \Big( B_i \setminus  \big(\bigcup_{j=1}^{N_i}C^{i,j}\big) \Big) \cup
\Big(  \bigcup_{l=1}^{i-1} \big( B_i \cap B_l' \big) \Big).
\end{align*}
Since $B_i\cap B_l=\emptyset$ for every $l=1,2,\dots,i-1,$ we have
$B_i\cap B_l' \subseteq B_l' \setminus B_l.$ Therefore,
$$
B_i \setminus B_i' \subseteq
\Big( B_i \setminus  \big(\bigcup_{j=1}^{N_i}C^{i,j}\big) \Big) \cup
\Big(  \bigcup_{l=1}^{i-1} \big( B_l' \setminus B_l \big) \Big)
\subseteq
\Big( B_i'' \setminus  \big(\bigcup_{j=1}^{N_i}C^{i,j}\big) \Big) \cup
\Big(  \bigcup_{l=1}^{i-1} \big( B_l'' \setminus B_l \big) \Big).
$$
As a result, for any $\underline{j}\in[k]^n$
\begin{equation} \label{eq:otherside}
\begin{aligned}
  P_{\underline{j}}(B_i) 
- P_{\underline{j}} (B_i') 
&\le  P_{\underline{j}}(B_i\setminus B_i') \\
&\le   P_{\underline{j}}\Big( B_i'' \setminus  \Big(\bigcup_{j=1}^{N_i}C^{i,j}\Big) \Big) 
+\sum_{l=1}^{i-1} P_{\underline{j}}(B_l'' \setminus B_l) \\
&\overset{(a)}{<}  \frac{i \alpha}{N^2} <\frac{\alpha}{N}
,
\end{aligned}
\end{equation}
where $(a)$ follows from \eqref{eq:setdiff}, \eqref{eq:Ni}, and the fact that $ \big(\bigcup_{j=1}^{N_i}C^{i,j}\big) \subseteq B_i''.$
Combining \eqref{eq:oneside} and \eqref{eq:otherside}, we obtain that for any $\underline{j}\in[k]^n$ and $i=2,3,\dots,N-1,$
$$
| P_{\underline{j}}(B_i') - P_{\underline{j}} (B_i) |< \frac{\alpha}{N} 
$$
Since both $\{B_i\}_{i=1}^N$ and $\{B_i'\}_{i=1}^N$ are partitions of $\cY^n,$ we have
$$
\sum_{i=1}^N P_{\underline{j}}(B_i) 
= \sum_{i=1}^N P_{\underline{j}}(B_i') = 1
\text{~for all~}\underline{j}\in [n]^k.
$$
As a result,
\begin{align*}
 | P_{\underline{j}}(B_N') 
- P_{\underline{j}} (B_N) | &= \left| \sum_{i=1}^{N-1} P_{\underline{j}}(B_i') 
- \sum_{i=1}^{N-1} P_{\underline{j}} (B_i) \right| \\
&\le \sum_{i=1}^{N-1} | P_{\underline{j}}(B_i') 
- P_{\underline{j}} (B_i) |
< \frac{(N-1)\alpha}{N} < \alpha.
\end{align*}
This shows that the sets $\{B_i'\}_{i=1}^N$ satisfy \eqref{eq:alpha} in the claim of the lemma.

(c) {\em Existence of the sets $A_i, i=1,\dots, L.$}
We define a finite collection of sets $\cC=\bigcup_{i=1}^{N-1}\{C^{i,j}:1\le j \le N_i\}.$ According to definition \eqref{eq:defBi}, $B_i'\in \sigma(\cC)$ for all $i=1,2,\dots,N,$ where $\sigma(\cC)$ is the finite algebra generated by $\cC.$
Since $\cC\subseteq \cR,$ for every $C^{i,j}\in \cC,$ we can write 
$
C^{i,j}=C_1^{i,j}\times C_2^{i,j}\times \dots \times C_n^{i,j},
$
where $C_l^{i,j}\in \sigma(\cY)$ for all $l=1,2,\dots,n.$
We define another finite collection of sets $\cE=\bigcup_{l=1}^n \bigcup_{i=1}^{N-1}\{C_l^{i,j}:1\le j \le N_i\}.$ 
It is clear that $\cE\subseteq \sigma(\cY).$ 
Let $\sigma_F(\cY) = \sigma(\cE),$ where $\sigma(\cE)$ is the finite algebra generated by $\cE.$ 
Then $\sigma_F(\cY) \subseteq \sigma(\cY).$ Moreover, 
the set of $n$-dimensional measurable rectangles with respect to $\sigma_F(\cY)$ is
$$
\cR_F=\{C_1\times C_2\times \dots \times C_n: C_i\in \sigma_F(\cY)\text{ for all } i=1,2,\dots,n\},
$$
and the $n$-fold finite product algebra 
$\sigma_F(\cY)\times \sigma_F(\cY)\times\dots\times\sigma_F(\cY)$ is the finite algebra generated by $\cR_F,$ and can be simply written as $\sigma(\cR_F).$
Since $\cC\subseteq \cR_F,$ $\sigma(\cC) \subseteq \sigma(\cR_F).$ Consequently, $B_i'\in \sigma(\cR_F)$
for all $i=1,2,\dots,N.$

It is well known that finite algebras are generated by finite partitions \cite[Lemma~1.3. and Remark~1.4.]{Koralov07}. As a result, there is a finite partition $\{A_i\}_{i=1}^L$ of $\cY$ such that $\sigma_F(\cY)=\sigma(\{A_i\}_{i=1}^L).$ Since $\sigma_F(\cY) \subseteq \sigma(\cY),$ we deduce that 
$A_i\in\sigma(\cY)$ for all $i=1,2,\dots,L.$ The proof is complete.
\bibliographystyle{IEEEtran}
\bibliography{differential}

\end{document}